\documentclass[3p,twocolumn,times]{elsarticle}

\usepackage{ecrc-modified}

\usepackage{graphicx}
\usepackage{amsmath}
\usepackage{epstopdf}
\usepackage{subcaption}
\usepackage{hyperref}
\usepackage{float}
\usepackage{algorithm}
\usepackage{amssymb}
\usepackage{multirow}
\usepackage{url}
\usepackage{soul}
\usepackage{color}
\usepackage{xspace}
\usepackage{paralist}
\usepackage{breqn} %to break long formula, breqn won't break after commas and ordinary atoms
\usepackage{cite}
\usepackage{import}
\usepackage{alltt}
\usepackage[table]{xcolor}
\usepackage{fixltx2e}
\usepackage{mathtools}
\usepackage{algpseudocode}% http://ctan.org/pkg/algorithmicx
\usepackage{array}
\usepackage{enumitem}
\usepackage{ccicons}
\usepackage{fancyhdr}
\newcommand{\LineComment}[1]{\Statex \(\triangleright\) #1}

\newcommand{\Let}[2]{\State #1 $\gets$ #2}

\renewcommand{\cite}{\citep}

\def\SB#1{\textsubscript{{#1}}}
\def\U#1{\textunderscore{{#1}}}

\newcolumntype{L}[1]{>{\raggedright\let\newline\\\arraybackslash\hspace{0pt}}m{#1}}
\newcolumntype{C}[1]{>{\centering\let\newline\\\arraybackslash\hspace{0pt}}m{#1}}
\newcolumntype{R}[1]{>{\raggedleft\let\newline\\\arraybackslash\hspace{0pt}}m{#1}}

\volume{00}

\firstpage{1}

\journalname{Web Semantics: Science, Services and Agents on the World Wide Web}

\runauth{M. Taheriyan}

\jid{}

\jnltitlelogo{Journal of Web Semantics}

\usepackage{amssymb}
\biboptions{}

\usepackage[figuresright]{rotating}

\begin{document}

\thispagestyle{fancy}

\begin{frontmatter}



\dochead{}

%\CopyrightLine{2016}{\url{http://dx.doi.org/10.1016/j.websem.2015.12.003} \copyright~2016. This manuscript version is made available under the CC-BY-NC-ND 4.0 license \url{http://creativecommons.org/licenses/by-nc-nd/4.0}.}

%\title{Learning the Semantics of Structured Data Sources\tnoteref{copyright}}
%\tnotetext[copyright]{\textcopyright~2016. This manuscript version is made available under the CC-BY-NC-ND 4.0 license:  \url{http://creativecommons.org/licenses/by-nc-nd/4.0}. DOI: \url{http://dx.doi.org/10.1016/j.websem.2015.12.003}. } 

\title{Learning the Semantics of Structured Data Sources}

\fancyfoot[CE,CO]{\thepage}
%\lfoot{ \footnotesize DOI: \url{http://dx.doi.org/10.1016/j.websem.2015.12.003} \textcopyright~2016. This manuscript version is made available under the CC-BY-NC-ND 4.0 license:  \url{http://creativecommons.org/licenses/by-nc-nd/4.0}.  }
\fancyhead[CE,CO]{\footnotesize \textit{M. Taheriyan et al. / Web Semantics: Science, Services and Agents on the World Wide Web} \normalsize}
%\ccLogo \ccAttribution \ccNonCommercial \ccNoDerivatives 


%\author[a]{Mohsen Taheriyan}
\author{Mohsen Taheriyan\corref{cor1}}
\ead{mohsen@isi.edu}
\cortext[cor1]{Corresponding author}
%\fntext[fn1]{Student}
%\author[a]{Craig Knoblock}
\author{Craig A. Knoblock}
\ead{knoblock@isi.edu}
%\author[a]{Pedro Szekely}
\author{Pedro Szekely}
\ead{pszekely@isi.edu}
%\author[a]{Jos\'{e} Luis Ambite}
\author{Jos\'{e} Luis Ambite}
\ead{ambite@isi.edu}

%\medskip

%\address[a]{University of Southern California, Information Sciences Institute, 4676 Admiralty Way, Marina del Rey, CA 90292, USA}
\address{University of Southern California, Information Sciences Institute, 4676 Admiralty Way, Marina del Rey, CA 90292, USA}


\begin{abstract}

Information sources such as relational databases, spreadsheets, XML, JSON, and Web APIs contain a tremendous amount of structured data that can be leveraged to build and augment knowledge graphs. However, they rarely provide a semantic model to describe their contents. Semantic models of data sources represent the implicit meaning of the data by specifying the concepts and the relationships within the data. Such models are the key ingredients to automatically publish the data into knowledge graphs. Manually modeling the semantics of data sources requires significant effort and expertise, and although desirable, building these models automatically is a challenging problem. Most of the related work focuses on semantic annotation of the data fields (source attributes). However, constructing a semantic model that explicitly describes the relationships between the attributes in addition to their semantic types is critical.

We present a novel approach that exploits the knowledge from a domain ontology and the semantic models of previously modeled sources to automatically learn a rich semantic model for a new source. This model represents the semantics of the new source in terms of the concepts and relationships defined by the domain ontology. Given some sample data from the new source, we leverage the knowledge in the domain ontology and the known semantic models to construct a weighted graph that represents the space of plausible semantic models for the new source. Then, we compute the top \textit{k} candidate semantic models and suggest to the user a ranked list of the semantic models for the new source. The approach takes into account user corrections to learn more accurate semantic models on future data sources. Our evaluation shows that our method generates expressive semantic models for data sources and services with minimal user input. These precise models make it possible to automatically integrate the data across sources and provide rich support for source discovery and service composition. They also make it possible to automatically publish semantic data into knowledge graphs.
 
\end{abstract}


\end{frontmatter}

%\linenumbers

%\let\clearpage\relax
% prevents pagebreaks after each include, the other solution is to use \input instead of \include

\section{Introduction}
\label{sec:introduction}

Knowledge graphs have recently emerged as a rich and flexible representation of domain knowledge. Nodes in this graph represent the entities and edges show the relationships between the entities. Large companies such as Google and Microsoft employ knowledge graphs as a complement for their traditional search methods to enhance the search results with semantic-search information. Linked Open Data (LOD) is an ongoing effort in the Semantic Web community to build a massive public knowledge graph. The goal is to extend the Web by publishing various open datasets as RDF on the Web and then linking data items to other useful information from different data sources. With linked data, starting from a certain point in the graph, a person or machine can explore the graph to find other related data. The focus of this work is the first step of publishing linked data, automatically publishing datasets as RDF using a common domain ontology. 

\begin{figure}[th]
\begin{center}
\includegraphics[width=.95\linewidth]{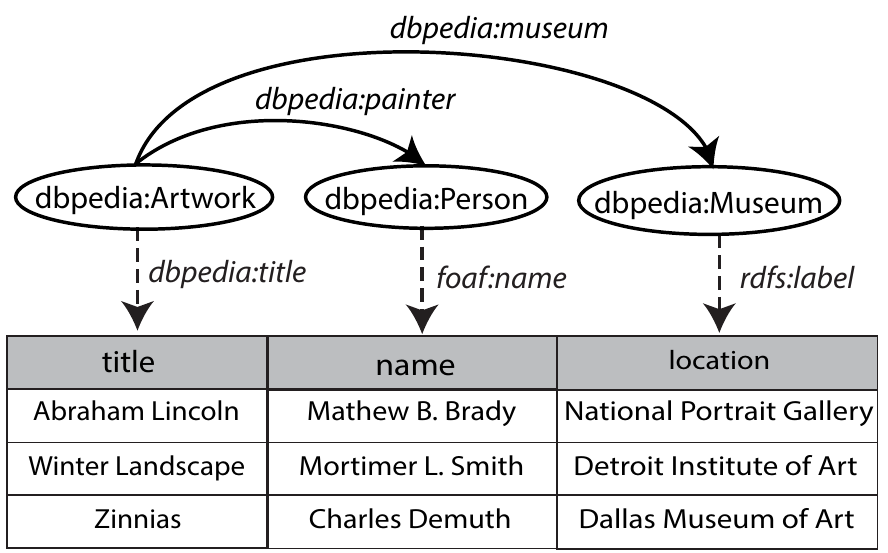}
\caption{The semantic model of a sample data source containing information about paintings.}
\label{fig:example}
\end{center}
\end{figure}

A large amount of data in LOD comes from structured sources such as relational databases and spreadsheets. Publishing these sources into LOD involves constructing \textit{source descriptions} that represent the intended meaning of the data by specifying mappings between the sources and the \textit{domain ontology} \cite{doan2012}. A domain ontology is a formal model that represents the concepts within a domain and the properties and interrelationships of those concepts. 
%Given all the structured data available, we would like to combine the data to answer specific user queries. A common approach to integrate sources involves building a domain model and constructing \textit{source descriptions} that represent the intended meaning of the data by specifying mappings between the sources and the domain model \cite{doan2012}.
%In the traditional data integration approaches, source descriptions are specified as global-as-view (GAV) or local-as-view (LAV) descriptions.
%In the Semantic Web, the domain model is an \textit{ontology} that formally represents concepts within a domain and properties and interrelationships of those concepts. 
In this context, what is meant by a source description is a schema mapping from the source to an ontology. We can represent this mapping as a \textit{semantic network} with ontology classes as the nodes and ontology properties as the links between the nodes. This network, also called a \textit{semantic model}, describes the source in terms of the concepts and relationships defined by the domain ontology. Figure~\ref{fig:example} depicts a semantic model for a sample data source including information about some paintings. This model explicitly represents the meaning of the data by mapping the source to the DBpedia\footnote{\url{http://dbpedia.org/ontology}} and FOAF\footnote{\url{http://xmlns.com/foaf/spec}} ontologies. Knowing this semantic model enables us to publish the data in the table into the LOD knowledge graph. 

% example: Person, Artwork, Place
%relationships: (Artwork,museum,Museum) - (Artwork,painter,Person)

% class hierarchy: Museum --> Building --> ArchitecturalStructure --> Place
% propert: location --> range: Place

One step in building a semantic model for a data source is \textit{semantic labeling}, determining the \textit{semantic types} of its data fields, or \textit{source attributes}. 
%\footnote{In this papers, the terms \textit{semantic type} and \textit{semantic label} are used interchangeably.}
That is, each source attribute is labeled with a class and/or a data property of the domain ontology. In our example in Figure~\ref{fig:example}, the semantic types of the first, second, and third columns are \textit{title} of \textit{Artwork}, \textit{name} of \textit{Person}, and \textit{label} of \textit{Museum} respectively. However, simply annotating the attributes is not sufficient. Unless the relationships between the columns are explicitly specified, we will not have a precise model of the data. In our example, a \textit{Person} could be the \textit{owner}, \textit{painter}, or \textit{sculptor} of an \textit{Artwork}, but in the context of the given source, only \textit{painter} correctly interprets the relationship between \textit{Artwork} and \textit{Person}. In the correct semantic model, \textit{Museum} is connected to \textit{Artwork} through the link \textit{museum}. Other models may connect \textit{Museum} to \textit{Person} instead of \textit{Artwork}. For instance, \textit{Person} could be \textit{president}, \textit{owner}, or \textit{founder} of \textit{Museum}, or \textit{Museum} could be \textit{employer} or \textit{workplace} of \textit{Person}. To build a semantic model that fully recovers the semantics of the data, we need a second step that determines the relationships between the source attributes in terms of the properties in the ontology.

%Writing source descriptions by hand is difficult, requiring significant effort and expertise. Although desirable, generating these descriptions automatically is a challenging problem. Most of the proposed approaches on the Semantic Web focus on the first step of the modeling process, mapping the source attributes to the concepts and data properties of an ontology. Nonetheless, the problem of how to get a precise semantic description of a data source that fully characterizes the information semantics still remains. 

Manually constructing semantic models requires significant effort and expertise. Although desirable, generating these models automatically is a challenging problem.
% \cite{alexe2011, dhamankar04, fagin2009:clio, marnette2011, rahm01}. 
In Semantic Web research, there is much work
% are many studies 
on mapping data sources to ontologies 
\cite{han2008:rdf123,sheth2008,langegger2009:xlwrap,sahoo2009:survey,polfliet2010:automated,limaye2010,vavliakis2010:rdote,ding2010:twc,saquicela2011,venetis2011,wang2012:understanding,mulwad2013semantic}, but most focus on semantic labeling or are very limited in automatically inferring the relationships. Our goal is to construct semantic models that not only include the semantic types of the source attributes, but also describe the relationships between them. 

In this paper, we present a novel approach that exploits the knowledge from a domain ontology and \textit{known semantic models} of sources in the same domain to automatically learn a rich semantic model for a new source. The work is inspired by the idea that different sources in the same domain often provide similar or overlapping data and have similar semantic models. Given sample data from the new source, we use a labeling technique \cite{krishnamurthy2015} to annotate each source attribute with a set of candidate semantic types from the ontology. 
%Next, we build a weighted directed graph with known semantic models as the main components and expand the graph by adding the paths in the ontology connecting the nodes in the graph. 
Next, we build a weighted directed graph from the known semantic models, learned semantic types, and the domain ontology. This graph models the space of plausible semantic models. Then, we find the most promising mappings from the source attributes to the nodes of the graph, and for each mapping, we generate a candidate model by computing the minimal tree that connects the mapped nodes. Finally, we score the candidate models to prefer the ones formed with more coherent and frequent patterns. 

This work builds on top of our previous work on learning semantic models of sources \cite{taheriyan14:icsc,taheriyan13:iswc}. The central data structure of our approach to learn a semantic model for a new source is a graph built on top of the known semantic models. In the previous work, we add a new component to the graph for each known semantic model. If two semantic models are very similar to each other and they only differ in one link, for example, we will still have two different components for them in the graph. The graph grows as the number of known semantic models grows, which makes computing the semantic models inefficient if we have a large set of known semantic models. In this paper, we extend our previous work to make it scale to a large number of semantic models. We present a new algorithm that constructs a much more compact graph by merging overlapping segments of the known semantic models. The new technique significantly reduces the size of the graph in terms of the number of nodes and links. Consequently, it considerably decreases the number of possible mappings from the source attributes to the nodes of the graph. It also makes computing the minimal tree that connects the nodes of the candidate mappings in the graph more efficient. The new method to build the graph changes our algorithms to compute and rank the candidate semantic models. 

%The main contribution of our work is a scalable approach to leverage attribute relationships in known semantic models to hypothesize attribute relationships for new sources, and capturing them in semantic models. 
The main contribution of this paper
% our work 
is a scalable approach that exploits the structure of the domain ontology and the known semantic models to build semantic models of new sources. 
We evaluated our approach on a set of museum data sources modeled using two well-known \textit{data models} in the cultural heritage domain: Europeana Data Model (EDM) \cite{hennicke:2011}, and CIDOC Conceptual Reference Model (CIDOC-CRM) \cite{doerr:2003}. A data model standardizes how to map the data elements in a domain to a set of domain ontologies. The evaluation shows that our approach automatically generates high-quality semantic models that would have required significant user effort to create manually. It also shows that the semantic models learned using both the domain ontology and the known models are approximately 70\% more accurate than the models learned with the domain ontology as the only background knowledge. The generated semantic models are the key ingredients to automate tasks such as source discovery, information integration, and service composition. They can also be formalized using mapping languages such as R2RML\cite{R2RML}, which can be used for converting data sources into RDF and publishing them into the Linked Open Data (LOD) cloud or any other knowledge graph. 

We have implemented our approach in Karma~\cite{knoblock12:semi}, our data modeling and integration framework.\footnote{\url{http://karma.isi.edu}} Users can import data from a variety of sources including relational databases, spreadsheet, XML files and JSON files into Karma. They can also import the domain ontologies they want to use for modeling the data. The system then automatically suggests a semantic model for the loaded source. Karma provides an easy to use graphical user interface to let users interactively refine the learned semantic models if needed. Once a semantic model is created for the new source, users can publish the data as RDF by clicking a single button. Szekely et al.~\cite{szekely2013:eswc} used Karma to model the data from Smithsonian American Art Museum\footnote{\url{http://americanart.si.edu}} and then publish it into the Linked Open Data cloud. Karma is also able to build semantic models for Web services and then exploits the created semantic models to build APIs that directly communicate at the semantic level \cite{szekely11:sso,taheriyan12:lapis,taheriyan12:rapidly}. 

\section{Motivating Example}
\label{sec:example}

\begin{figure}[tph]
\centering
\begin{subfigure}[b]{.48\textwidth}
  \centering
  %\imagebox{48mm}{\includegraphics[width=.95\linewidth]{figures/dma}}
  % imagebox aligns top of the two images that are next to each other
  \includegraphics[width=.95\linewidth]{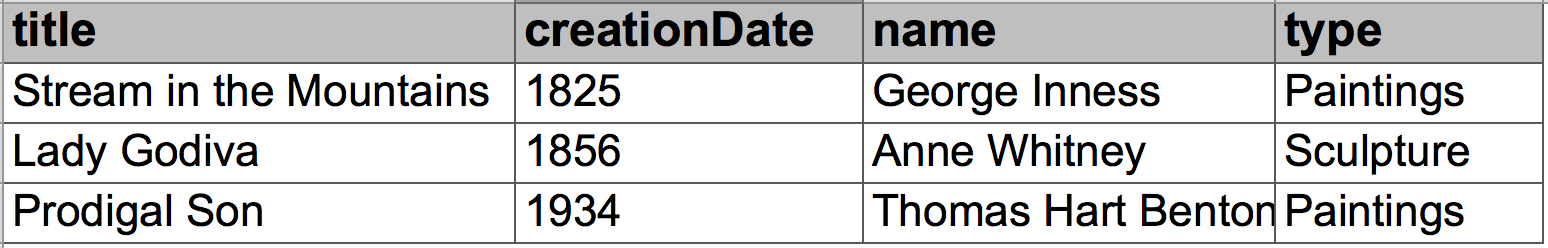}
  \caption{\textit{dma(title, creationDate, name, type)}}
  \vspace{0.05in}
  \label{fig:dma-data}
\end{subfigure} \\
% removing the new line (\\) from the end puts two images next to each other
\begin{subfigure}[b]{.48\textwidth}
  \centering
  %\imagebox{48mm}{\includegraphics[width=.95\linewidth]{figures/npg}}
  \includegraphics[width=.95\linewidth]{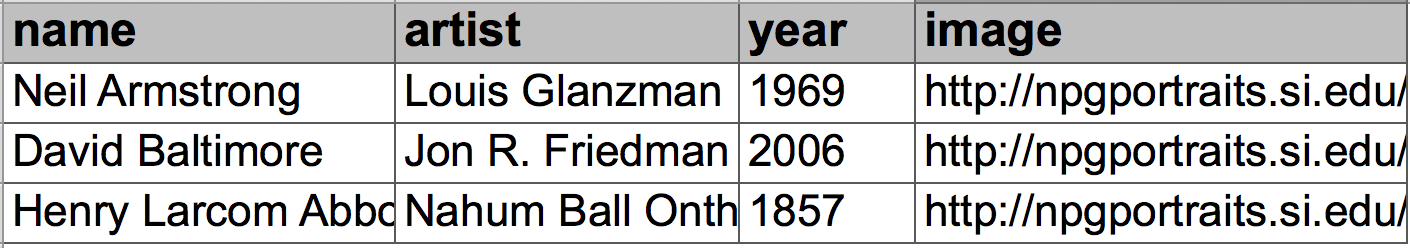}
  \caption{\textit{npg(name, artist, year, image)}}
  \vspace{0.05in}
  \label{fig:npg-data}
\end{subfigure}
\begin{subfigure}[b]{.48\textwidth}
  \centering
  %\imagebox{48mm}{\includegraphics[width=.95\linewidth]{figures/npg}}
  \includegraphics[width=.95\linewidth]{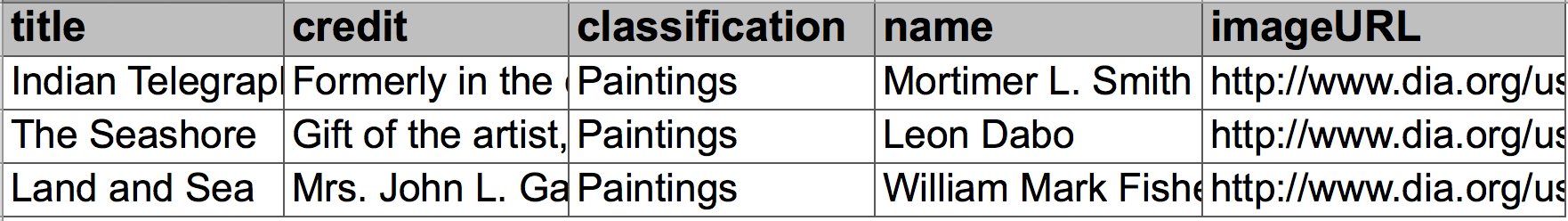}
  \caption{\textit{dia(title, credit, classification, name, imageURL)}}
  \vspace{0.05in}
  \label{fig:dia-data}
\end{subfigure}
\caption{Sample data from three museum sources: (a) Dallas Museum of Art, (b) National Portrait Gallery, and (c) Detroit Institute of Art.}
\label{fig:semantic-models}
\end{figure}

We explain the problem of learning semantic models by giving a concrete example that will be used throughout this paper to illustrate different steps of our approach. In this example, the goal is to model a set of museum data sources using EDM\footnote{\url{http://www.europeana.eu/schemas/edm}}, AAC\footnote{\url{http://www.americanartcollaborative.org/ontology}}, SKOS\footnote{\url{http://www.w3.org/2008/05/skos\#}}, Dublin Core Metadata Terms\footnote{\url{http://purl.org/dc/terms}}, FRBR\footnote{\url{http://vocab.org/frbr/core.html}}, FOAF%\footnote{\url{http://xmlns.com/foaf/spec}}
, ORE\footnote{\url{http://www.openarchives.org/ore/terms}}, and ElementsGr2\footnote{\url{http://rdvocab.info/ElementsGr2}} ontologies and then use the created semantic models to publish their data as RDF \cite{szekely2013:eswc}. Suppose that we have three data sources. The first source is a table containing information about artworks in the Dallas Museum of Art\footnote{\url{http://www.dma.org}} (Figure~\ref{fig:dma-data}). We formally write the signature of this source as \textit{dma(title, creationDate, name, type)} where \textit{dma} is the name of the source and \textit{title}, \textit{creationDate}, \textit{name}, and \textit{type} are the names of the source attributes (columns). The second source, \textit{npg}, is a CSV file including the data of some of the portraits in the National Portrait Gallery\footnote{\url{http://www.nationalportraitgallery.org}} (Figure~\ref{fig:npg-data}), and the third data source, \textit{dia}, has the data of the artworks in the Detroit Institute of Art\footnote{\url{http://www.dia.org}} (Figure~\ref{fig:dia-data}). 

%% \usepackage{subgig}
%\begin{figure}[tbph]
%    \centering
%    \subfloat[]{{\includegraphics[width=.4\linewidth]{figures/dma} }}
%    \qquad
%    \subfloat[]{{\includegraphics[width=.4\linewidth]{figures/npg} }}%
%    \caption{2 Figures side by side}%
%    \label{fig:known-models}%
%\end{figure}

%\begin{figure}[t]
%\centering
%\begin{subfigure}[b]{.5\textwidth}
%  \centering
%  %\imagebox{48mm}{\includegraphics[width=.95\linewidth]{figures/dma}}
%  % imagebox aligns top of the two images that are next to each other
%  \includegraphics[width=.95\linewidth]{figures/dma}
%  \caption{}
%  \label{fig:dma}
%\end{subfigure} \\
%% removing the new line (\\) from the end puts two images next to each other
%\begin{subfigure}[b]{.5\textwidth}
%  \centering
%  %\imagebox{48mm}{\includegraphics[width=.95\linewidth]{figures/npg}}
%  \includegraphics[width=.95\linewidth]{figures/npg}
%  \caption{}
%  \label{fig:npg}
%\end{subfigure}
%\caption{Known semantic models: (a) semantic model of the source \textit{dma}, and (b) semantic model of the source \textit{npg}. Class nodes (ovals) and links are labeled with URIs (prefixed by the ontology namespace). The link \textit{uri} in (b) simply means that the values of the attribute \textit{image} in the source \textit{npg} are URIs of instances of the class \textit{edm:WebResource}. }
%\label{fig:known-models}
%\end{figure}
%
%\begin{figure}[t]
%\begin{center}
%%\scalebox{0.7} {\includegraphics[width=4.4in]{figures/dia}}
%\includegraphics[width=.95\linewidth]{figures/dia}
%\caption{Semantic model of the new source \textit{dia}}
%\label{fig:dia}
%\end{center}
%\end{figure}

\begin{figure}[tph]
\centering
\begin{subfigure}[b]{.5\textwidth}
  \centering
  %\imagebox{48mm}{\includegraphics[width=.95\linewidth]{figures/dma}}
  % imagebox aligns top of the two images that are next to each other
  \includegraphics[width=.95\linewidth]{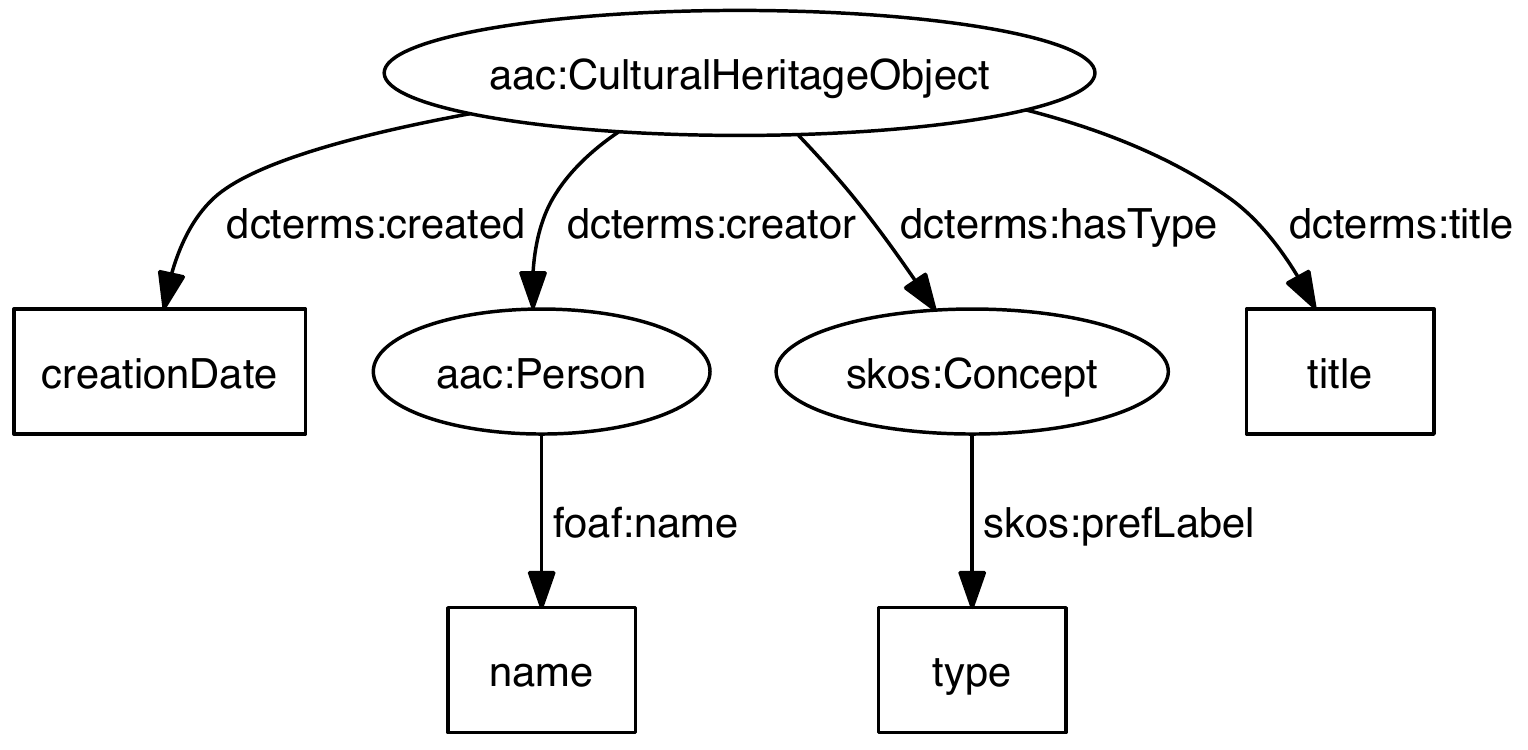}
  \caption{\textit{sm(dma)}: semantic model of the source \textit{dma}}
  \label{fig:dma}
  \vspace{0.2in}
\end{subfigure} \\
% removing the new line (\\) from the end puts two images next to each other
\begin{subfigure}[b]{.5\textwidth}
  \centering
  %\imagebox{48mm}{\includegraphics[width=.95\linewidth]{figures/npg}}
  \includegraphics[width=.95\linewidth]{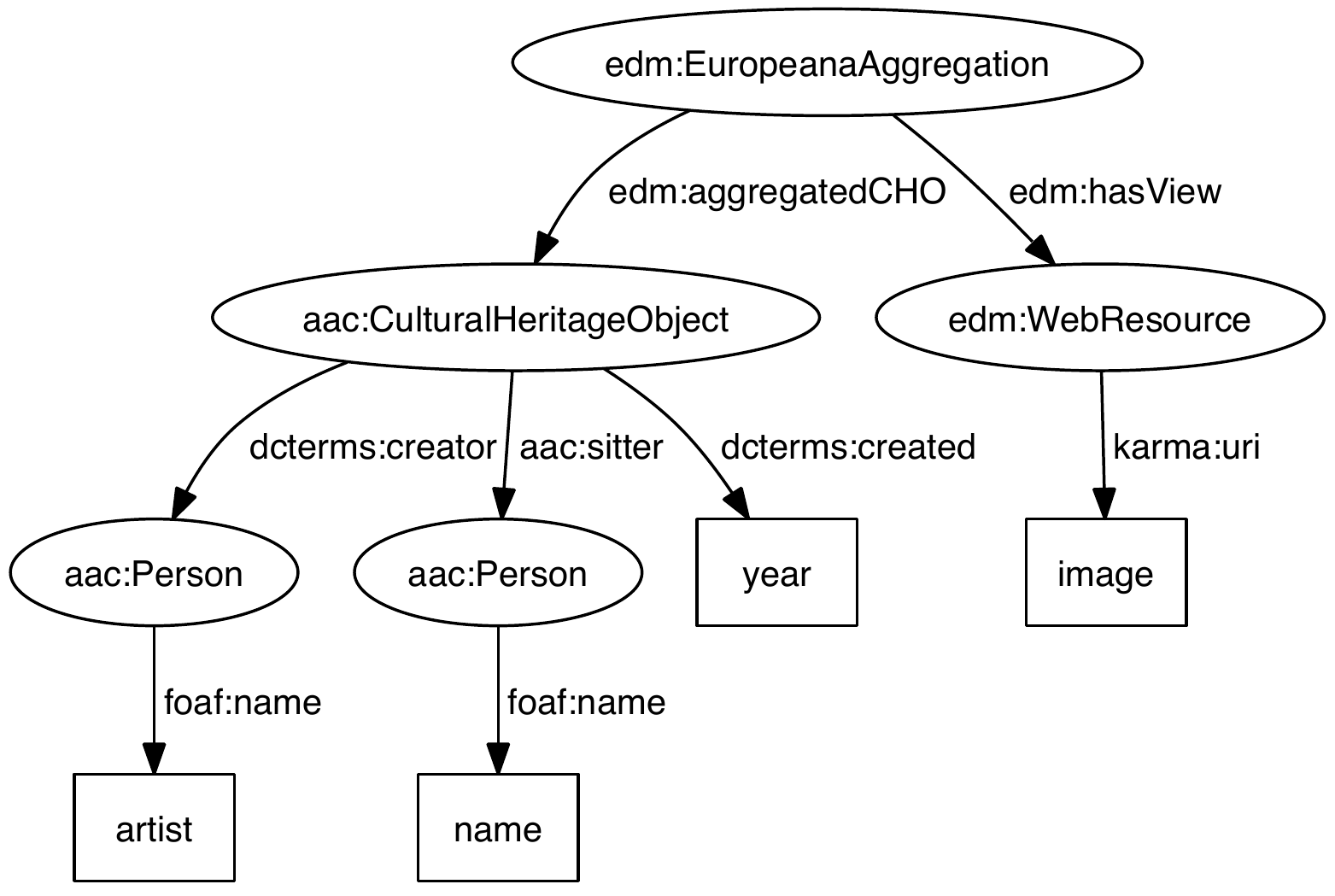}
  \caption{\textit{sm(npg)}: semantic model of the source \textit{npg}}
  \label{fig:npg}
  \vspace{0.2in}
\end{subfigure}
\begin{subfigure}[b]{.5\textwidth}
  \centering
  %\imagebox{48mm}{\includegraphics[width=.95\linewidth]{figures/npg}}
  \includegraphics[width=.95\linewidth]{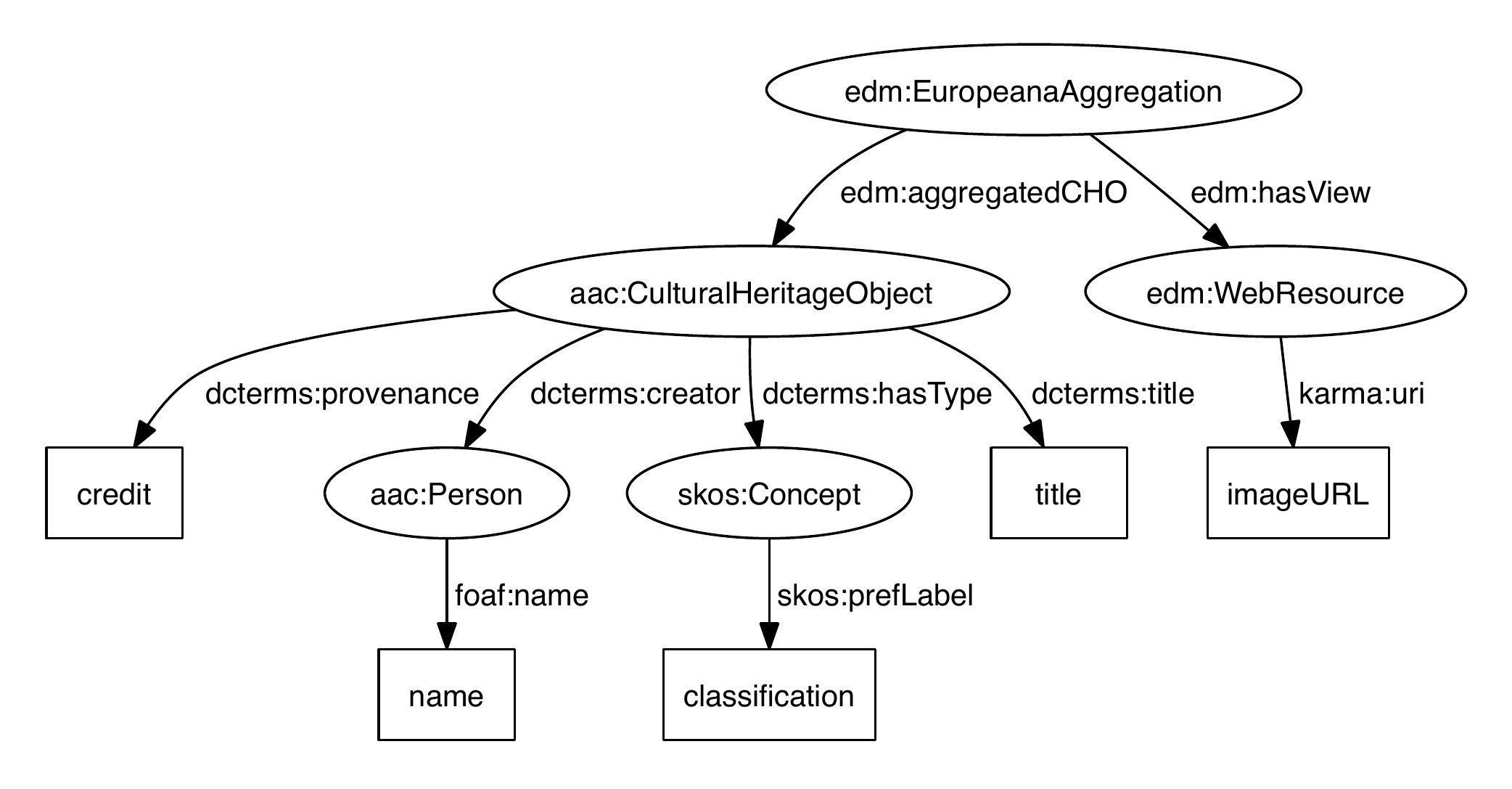}
  \caption{\textit{sm(dia)}: semantic model of the source \textit{dia}}
  \label{fig:dia}
  \vspace{0.2in}
\end{subfigure}
\caption{Semantic models of the example data sources created by experts in the museum domain. Class nodes (ovals) and links correspond to classes and properties in the ontology (prefixed by the ontology namespace). The particular link \textit{karma:uri} in (b) simply means that the values of the attribute \textit{image} in the source \textit{npg} are the URIs of the instances of the class \textit{edm:WebResource}. }
\label{fig:semantic-models}
\end{figure}

Figure~\ref{fig:semantic-models} shows the correct semantic model of the sources \textit{dma}, \textit{npg}, and \textit{dia} created by experts in the museum domain. A semantic model of the source \textit{s}, called \textit{sm(s)}, is a directed graph containing two types of nodes. \textit{Class nodes} (ovals) correspond to classes in the ontology, and 
% ($v \in class~nodes \Rightarrow l_v=class\_uri$). 
\textit{data nodes} (rectangles) correspond to the source attributes (labeled with the attribute names). 
%($v \in data~nodes \Rightarrow l_v=attribute\_name$). 
The links in the graph are associated with ontology properties. 
%($e \in links \Rightarrow l_e=property\_uri$). 
%The particular link \textit{karma:uri} from a class node to a data node denotes that the values of the attribute corresponding to that data node are URIs of the instances of the class specified by that class node. 
The particular link \textit{karma:uri} from a class node, which represents an ontology class, to a data node, which represents a source attribute,  denotes that the attribute values are the URIs of the class instances.
For instance, in Figure~\ref{fig:npg}, the values of the column \textit{image} in the source \textit{npg} are the URIs of the instances of the class \textit{edm:WebResource}.

As discussed earlier, automatically building the semantic models is difficult. Machine learning methods can help us in assigning semantic types to the attributes by looking into the attributes values, however, these methods are error prone when similar data values have different semantic types. For example, from just the data values of the attribute \textit{creationDate} in the source \textit{dma}, it is hard to say whether it is the creation date of \textit{aac:CulturalHeritageObject} or it is the birthdate of a \textit{aac:Person}. Extracting the relationships between the attributes is a more complicated problem. There might be multiple paths connecting two classes in the ontology and we do not know which one captures the intended meaning of the data. For instance, there are several paths in the domain ontology connecting \textit{aac:CulturalHeritageObject} to \textit{aac:Person}, but in the context of the source \textit{dma}, only the link \textit{dcterms:creator} represents the correct meaning of the source. As another example, the attributes \textit{artist} and \textit{name} in the source \textit{npg} are both labeled with name of \textit{Person}, nevertheless, how can we decide whether these two attributes are different names of one person or they belong to two distinct individuals? In general, the ontology defines a large space of possible semantic models and without additional context, we do not know which one describes the source more precisely. 

%Some work \cite{knoblock12:semi} takes user inputs to reduce the size of the search range and semi-automatically maps a new source to a an ontology. 

Now, assume that the correct semantic models of the sources \textit{dma} and \textit{npg} are given. Can we leverage these known semantic models to build a semantic model for a new source such as \textit{dia}? In the next section, we present a scalable and automated approach that exploits the known semantic models \textit{sm(dma)} and \textit{sm(npg)} to limit the search space and learn a semantic model \textit{sm(dia)} for the new source \textit{dia}.

\section{Learning Semantic Models}
\label{sec:learning}

We now formally state the problem of learning semantic models of data sources. Let $O$ be the domain ontology\footnote{$O$ can be a set of ontologies.} and $\{sm(s_1),\allowbreak sm(s_2),\allowbreak \cdots,\allowbreak sm(s_n)\}$ is a set of known semantic models corresponding to the data sources $\{s_1,\allowbreak s_2,\allowbreak \cdots,\allowbreak s_n\}$. Given sample data from a new source $s(a_1,\allowbreak a_2,\allowbreak \cdots,\allowbreak a_m)$ called the \textit{target source}, in which $\{a_1,\allowbreak a_2,\allowbreak \cdots,\allowbreak a_m\}$ are the source attributes, our goal is to automatically compute a semantic model $sm(s)$ that captures the intended meaning of the source $s$.   In our example, \textit{sm(dma)} and \textit{sm(npg)} are the known semantic models, and the source \textit{dia} is the new source for which we want to automatically learn a semantic model.

The main idea is that data sources in the same domain usually provide overlapping data. Therefore, we can leverage attribute relationships in known semantic models to hypothesize attribute relationships for new sources. One of the metrics helping us to infer relationships between the attributes of a new source is the popularity of the links between the semantic types in the set of known models. Nevertheless, simply using link popularity to connect a set of nodes would lead to myopic decisions that select links that appear frequently in other models without taking into account how these nodes are connected to other nodes in the given models. Suppose that we have a set of 5 known semantic models. One of these models contains the link \textit{painter} between \textit{Artwork} and \textit{Person} and the link \textit{museum} between \textit{Artwork} and \textit{Museum} (similar to the example in Figure~\ref{fig:example}). The other 4 models do not contain the type \textit{Artwork}, but they include the link \textit{founder} from \textit{Museum} to \textit{Person}. If a given new source contains the types \textit{Artwork}, \textit{Museum}, and \textit{Person}, just using the link popularity yields to an incorrect model. Our approach takes into account the coherence of the patterns in addition to their popularity, and this is more complicated to do. 
 
Our approach to learn a semantic model for a new source has four steps: 
(1) Using sample data from the new source, learn the semantic types of the source attributes.  
(2) Construct a graph from the known semantic models, augmented with nodes and links corresponding to the learned semantic types and ontology paths connecting nodes of the graph.
(3) Compute the candidate mappings from the source attributes to the nodes of the graph. 
(4) Finally, build candidate semantic models for the candidate mappings, and rank the generated models.
\subsection{Learning Semantic Types of Source Attributes}
\label{sec:labeling}

The first step to model the semantics of a new source is to recognize the semantic types of its data. We call this step semantic labeling, which involves annotating the source columns with classes or properties in an ontology. The objective of this step is to assign semantic types to source attributes. We formally define a semantic type to be either an ontology class \textit{$\langle$class\U uri$\rangle$} or a pair consisting of a domain class and one of its data properties \textit{$\langle$class\U uri,property\U uri$\rangle$}. We use a class as a semantic type for attributes whose values are URIs for instances of a class and for attributes containing automatically-generated database keys that can also be modeled as instances of a class. We use a domain/data property pair as a semantic type for attributes containing literal values. For example, the semantic types of the attributes \textit{imageURL} and \textit{classification} in the source \textit{dia} are respectively \textit{$\langle$edm:WebResource$\rangle$} and \textit{$\langle$skos:Concept,skos:prefLabel$\rangle$}. 

While syntactic information about data sources such as attribute names or attribute types (string, int, date, ...) may give the system some hints to discover semantic types, they are often not sufficient, e.g., name of the first field in the source \textit{dia} is \textit{title} and we do not know whether this is a title of a book, song, or an artwork. Moreover, in many cases, attribute names are used in abbreviated forms, e.g., \textit{dob} rather than \textit{birthdate}. 

We employ the technique proposed by Krishnamurthy et al. \cite{krishnamurthy2015} to learn semantic types of source attributes. Their approach focuses on learning the semantic types from the data rather than the attribute names. It learns a semantic labeling function from a set of sources that have been manually labeled. When presented with a new source, the learned semantic labeling function can automatically assign semantic types to each attribute of the new source. The training data consists of a set of semantic types and each semantic type has a set of data values and attribute names associated with it. Given a new set of data values from a new source, the goal is to predict the top \textit{k} candidate semantic types along with confidence scores using the training data. 

If the data values associated with a source attribute $a_i$ are textual data, the labeling algorithm uses the \textit{cosine similarity} between TF/IDF vectors of the labeled documents and the input document to predict candidate semantic types. The set of data values associated with each textual semantic type in the training data is treated as a document, and the input document consists of the data values associated with $a_i$. For attributes with numeric data, the algorithm uses \textit{statistical hypothesis testing} \cite{lehmann2005:testing} to analyze the distribution of numeric values. The intuition is that the distribution of values in each semantic type is different. For example, the distribution of temperatures is likely to be different from the distribution of weights. The training data here consists of a set of numeric semantic types and each semantic type has a sample of numeric data values. At prediction time, given a new set of numeric data values (query sample), the algorithm performs statistical hypothesis tests between the query sample and each sample in the training data.

Once we apply this labeling method, it generates a set of candidate semantic types for each source attribute, each with a confidence value. Our algorithm then selects the top \textit{k} semantic types for each attribute as an input to the next step of the process. Thus, the output of the labeling step for $s(a_1,\allowbreak a_2,\allowbreak \cdots,\allowbreak a_m)$ is 
$T=\{(t_{11}^{p_{11}},\allowbreak \cdots,\allowbreak t_{1k}^{p_{1k}}),\allowbreak \cdots,\allowbreak (t_{m1}^{p_{m1}},\allowbreak \cdots,\allowbreak t_{mk}^{p_{mk}})\}$, 
where in $t_{ij}^{p_{ij}}$, $t_{ij}$ is the $j$th semantic type learned for the attribute $a_i$ and $p_{ij}$ is the associated confidence value which is a decimal value between 0 and 1. Table~\ref{tab:semantic-types} lists the candidate semantic types for the source \textit{dia} considering \textit{k=2}. 

\begin{table}[t]
\caption{Top two learned semantic types for the attributes of the source \textit{dia}.}
\vspace{-0.1in}
\begin{center}
\scalebox{1} {
\begin{tabular}{ |>{\centering\arraybackslash\itshape}m{1.8cm}| m{4.8cm} | }
\hline
\bf{attribute}  & \bf{candidate semantic types} \\ \hline

\multirow{2}{*} \textit{title} & \textit{$\langle$aac:CulturalHeritageObject,\newline dcterms:title$\rangle ^{0.49}$} \\
%\cline{2-2}
 & \textit{$\langle$aac:CulturalHeritageObject,\newline rdfs:label$\rangle ^{0.28}$} \\ \hline

\multirow{2}{*} \textit{credit} & \textit{$\langle$aac:CulturalHeritageObject,\newline dcterms:provenance$\rangle ^{0.83}$} \\
 & \textit{$\langle$aac:Person,ElementsGr2:note$\rangle^{0.06}$} \\ \hline

\multirow{2}{*} \textit{classification} & \textit{$\langle$skos:Concept,skos:prefLabel$\rangle^{0.58}$} \\
 & \textit{$\langle$skos:Concept,rdfs:label$\rangle^{0.41}$} \\ \hline

\multirow{2}{*} \textit{name} & \textit{$\langle$aac:Person,foaf:name$\rangle^{0.65}$} \\
 & \textit{$\langle$foaf:Person,foaf:name$\rangle^{0.32}$} \\ \hline

\multirow{2}{*} \textit{imageURL} & \textit{$\langle$foaf:Document$\rangle^{0.47}$} \\
 & \textit{$\langle$edm:WebResource$\rangle^{0.40}$} \\ \hline

\end{tabular}
}
\end{center}
\label{tab:semantic-types}
\vspace{-0.2in}
\end{table}

As we can see in Table~\ref{tab:semantic-types}, the semantic labeling method prefers \textit{$\langle$foaf:Document$\rangle$} for the semantic type of the attribute \textit{imageURL}, while according to the correct model (Figure~\ref{fig:dia}), \textit{$\langle$edm:WebResource$\rangle$} is the correct semantic type. We will show later how our approach recovers the correct semantic type by considering coherence of structure in computing the semantic models. 

\subsection{Building A Graph from Known Semantic Models, Semantic Types, and Domain Ontology}
\label{sec:graph}

So far, we have tagged the attributes of \textit{dia} with a set of candidate semantic types. To build a complete semantic model we still need to determine the relationships between the attributes. We leverage the knowledge of the known semantic models to discover the most popular and coherent patterns connecting the candidate semantic types. 

The central component of our method is a directed weighted graph $G$ built on top of the known semantic models and expanded using the semantic types $T$ and the domain ontology $O$. Similar to a semantic model, $G$ contains both class nodes and data nodes and links. The links correspond to properties in $O$ and there are weights on the links. Algorithm~\ref{alg:graph} shows the steps to build the graph. Our algorithm has three parts: (1) adding the known semantic models, $sm(dma)$ and $sm(npg)$ (Algorithm~\ref{alg:addknownmodels}); (2) adding the semantic types learned for the target source (Algorithm~\ref{alg:addsemantictypes}); and (3) expanding the graph using the domain ontology $O$ (Algorithm~\ref{alg:addontologypaths}). \\
%The algorithm we use to construct $G$ is described in detail in \hl{Appendix}. 
%Here, we provide an informal explanation of our algorithm to build the graph. \\

%\setlength{\textfloatsep}{0pt}% Remove \textfloatsep
\begin{algorithm}[t]
	\caption{\textit{Construct Graph G=(V,E)}}
	\label{alg:graph}
	{\fontsize{8}{8}\selectfont
	\begin{algorithmic}[1]
		\Statex \textbf{Input:} 
		\Statex ~~~~~~~~ - Known Semantic Models $M=\{sm_1, \cdots, sm_n\}$, 
		\Statex ~~~~~~~~ - Attributes(s) $A=\{a_1, \cdots, a_m\}$
		\Statex ~~~~~~~~ - Semantic Types $T=\{(t_{11}^{p_{11}}$, $\cdots$, $t_{1k}^{p_{1k}})$, $\cdots$, $(t_{m1}^{p_{m1}}$, $\cdots$, $t_{mk}^{p_{mk}})\}$
		\Statex ~~~~~~~~ - Ontology $O$
		\Statex \textbf{Output:} Graph $G=(V,E)$
		\newline
		\State \Call{AddKnownModels}{$G$,$M$}
		\State \Call{AddSemanticTypes}{$G$,$T$}
		\State \Call{AddOntologyPaths}{$G$,$O$}
		\newline \newline
		\Return $G$
		\newline
	\end{algorithmic}}
\end{algorithm}

\begin{figure*}[tph]
\begin{center}
\scalebox{1} {\includegraphics[width=.85\linewidth]{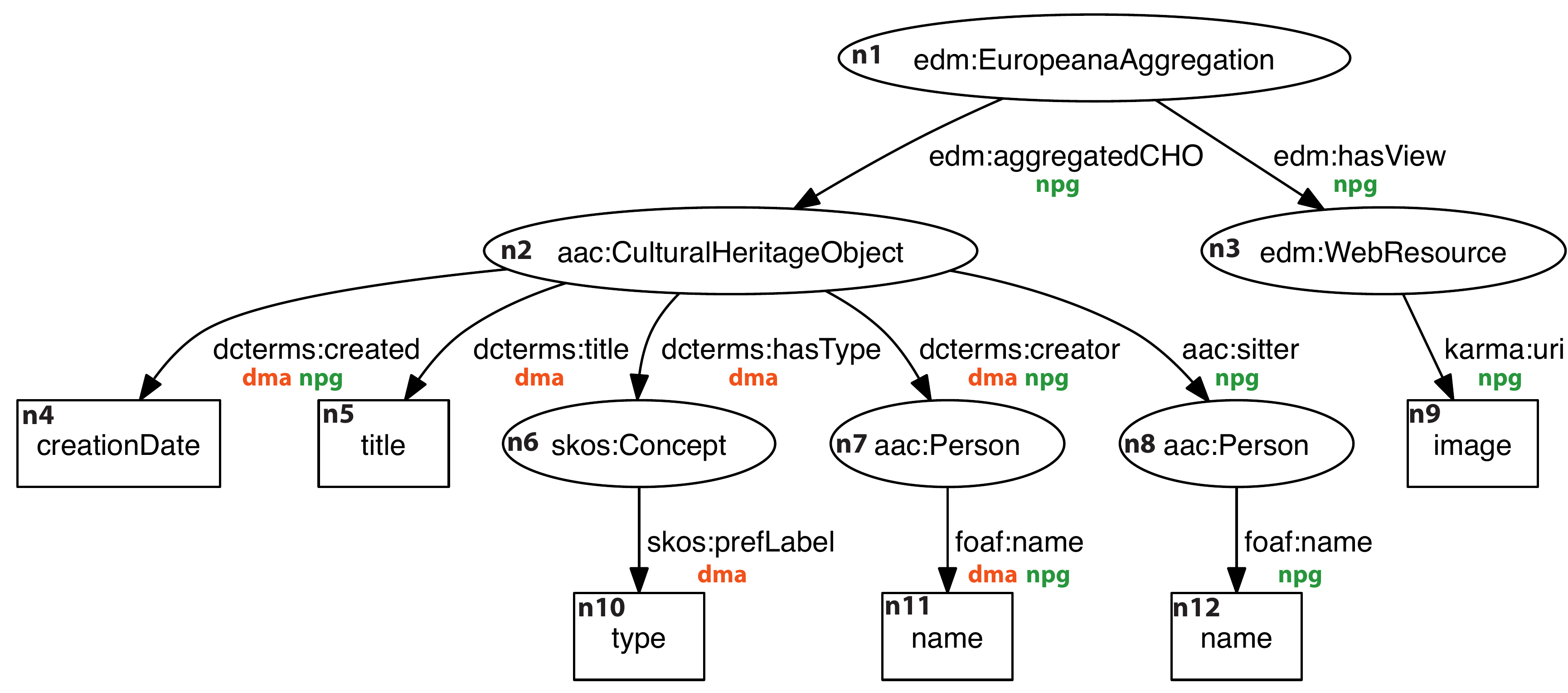}}
\caption{The graph $G$ after adding the known semantic models $sm(dma)$ and $sm(npg)$.}
\label{fig:graph-step1}
\end{center}
\end{figure*}

\noindent \textbf{Adding Known Semantic Models}: Suppose that we want to add $sm(s_i)$ to the graph. If the graph is empty, we simply add all the nodes and links in $sm(s_i)$ to $G$, otherwise we merge the nodes and links of $sm(s_i)$ into $G$ by adding the nodes and links that do not exist in $G$. When adding a new node or link, we tag it with a unique identifier (e.g., $s_i$, name of the source) indicating that the node/link exist in $sm(s_i)$. If a node or link already exists in the graph, we just add the identifier $s_i$ to its tags. The nodes and the links that are added in this step are shown with the black color in Figure~\ref{fig:graph-step1}. 
In order to easily refer to the nodes of the figure in the text, we assign a unique name to each node. The name of a node is written with small font at the left side of the node. For example, the node with the label \textit{edm:EuropeanaAggregartion} is named $n_1$. The orange and green tags below the labels of the black links are the identifiers indicating the semantic model(s) supporting the links. For instance, the link \textit{dcterms:creator} from $n_{2}$ (\textit{aac:CulturalHeritageObject}) to $n_{7}$ (\textit{aac:Person}) is tagged with both \textit{dma} and \textit{npg}, because it exists in both $sm(dma)$ and $sm(npg)$. For readability, we have not put the tags of the nodes in Figure~\ref{fig:graph-step1}.

Although merging a semantic model into $G$ looks straightforward, there are difficulties when the semantic model or the graph include multiple class nodes with the same label. Suppose that $G$ already includes two class nodes $v_1$ and $v_2$ both labeled with \textit{Person} connected by the link \textit{isFriendOf}. Now, we want to add a semantic model including the link \textit{worksFor} from \textit{Person} to \textit{Organization}. Assuming $G$ does not have a class node with the label \textit{Organization}, we add a new class node $v_3$ to $G$. Now, the question is where to put the link \textit{worksFor}, between $v_1$ and $v_3$, or $v_2$ and $v_3$. One option is to duplicate the link by adding a link between each pair and then assign different tags to the added links. This approach slows down the process of building the graph, and because it can yield a graph with a large number of links, our algorithm to compute the candidate semantic models would be inefficient too. Therefore, we adopt a different strategy; if there is more than one node in the graph matching a node in the semantic model, we select the one having more tags. This heuristic creates a more compact graph and makes the whole algorithm faster, while not having much impact on the results.
%In these cases, we duplicate the links, i.e., we add the link between all the matching pairs. 

%\setlength{\textfloatsep}{0pt}% Remove \textfloatsep
\begin{algorithm}[t]
	\caption{\textit{Add Known Semantic Models to G}}
	\label{alg:addknownmodels}
	{\fontsize{8}{8}\selectfont
	\begin{algorithmic}[1]
	
	\Function{AddKnownModels}{$G$,$M$}
		\newline
		\State{$\mathit{H=HashMap\langle node, node\rangle}$}
		\LineComment{H keys: nodes in $sm_i$}
		\LineComment{H values: matched nodes in $G$}
		\newline
    	%\LineComment{adding $sm_i$ to the graph}
    	\For  {each $sm_i \in M$} 
	    	\For {each class node $v$ in $sm_i$} 
		    	\Let{$l_v$}{label of $v$}
	    		\Let{$c_1$}{number of class nodes in $sm_i$ with label $l_v$}
	    		\Let{$c_2$}{number of class nodes in $G$ with label $l_v$}
	    		\State add $(c_1 - c_2)$ class nodes with label $l_v$ to $G$
	    		\Let{$\mathit{matched\_nodes}$}{class nodes with label $l_v$ in $G$}
	    		\Let{$\mathit{unmapped\_nodes}$}{$\mathit{matched\_nodes}-values(H)$}
	    		\Let{$v'$}{the node with largest tag set in $\mathit{unmapped\_nodes}$}
	    		\State add $\langle v,v'\rangle$ to $H$
    		\EndFor
    		\newline
	    	\For {each link $e$ from a class node $u$ to a \textbf{data} node $v$ in $sm_i$} 
		    	\Let{$l_e$}{label of $e$}
		    	\Let{$u'$}{$H(u)$}
				\If {$u'$ has an outgoing link with label $l_e$} 
					\Let{$v'$}{target of the link with label $l_e$}
				\Else
					\State add a new data node $v'$ to $G$
				\EndIf
				\State add $\langle v,v'\rangle$ to $H$
    		\EndFor
    		\newline
    		\Let{$w_l$}{$1$}
	    	\For {each link $e$ from $u$ to $v$ in $sm_i$} 
	    		\Let{$u'$}{$H(u)$}
	    		\Let{$v'$}{$H(v)$}
    			\If {there is $e'$ from $u'$ to $v'$ with $l_{e'}=l_{e}$ in $G$}
    				\Let{$\mathit{tags}_{e'}$}{$\mathit{tags}_{e'} \cup sm_{i}$} 
    				\Let{$weight(e')$}{$w_l-|\mathit{tags}_{e'}|/(i+1)$}
	   			\Else
	   				\State add the link $e'$ from $u'$ to $v'$ with $l_{e'}=l_e$ to $G$
    				\Let{$\mathit{tags}_{e'}$}{$sm_{i}$}
	   				\Let{$weight(e')$}{$w_l$}
	   			\EndIf
    		\EndFor
    	\EndFor
    	\newline
    \EndFunction	
	\end{algorithmic}}
\end{algorithm}

Algorithm~\ref{alg:addknownmodels} illustrates the details of our method to add a semantic model $sm(s_i)$ to $G$: 

\begin{enumerate}[leftmargin=*]

\item $[$line 2$]$ Let $H$ be a HashMap keeping the mappings from the nodes in $sm(s_i)$ to the nodes in $G$. The key of each entry in $H$ is a node in $sm(i)$ and its value is a node in $G$. 

\item $[$lines 3-13$]$: For each class node $v$ in $sm(s_i)$, we search the graph to see if $G$ includes a class node with the same label. If no such node exists in the graph, we simply add a new node to the graph. It is possible that $sm(s_i)$ contains multiple class nodes with the same label, for instance, a model including the link \textit{isFriendOf} from one \textit{Person} to another \textit{Person}. In this case, we make sure that $G$ also has at least the same number of class nodes with that label. For example, if $G$ only has one \textit{Person}, we add another class node with the label \textit{Person}. Once we added the required class nodes to the graph, we map the class nodes in the model to the class nodes in the graph. If $G$ has multiple class nodes with the same label, we select the one that is tagged by larger number of known semantic models. We add an entry to $H$ with $v$ as the key and the mapped node ($v'$) as the value.  

\item $[$lines 14-23$]$: For each link $e=(u,v)$ in $sm(s_i)$ where $v$ is a data node, we search the graph to see if there is a match for this pattern. We first use $H$ to find the node $u'$ in $G$ to which the node $u$ is mapped. If $u'$ does not have any outgoing link with a label equal to the label of $e$, we add a new data node $v'$ to $G$. We add $v$ and its mapped data node $v'$ to $H$. %In some cases, a class node $u$ in $sm(s_i)$ may be connected to multiple data nodes with the same data property (e.g., a model in which \textit{Person} has two links labeled with \textit{phone}). In this situation, we add more than one data node to the graph. For example, assume that there are two links $e_1=(u,v_1)$ and $e_2=(u,v_2)$ in $sm(s_i)$ and both links have the same label. When processing $e_1$, we add two data nodes to $G$ corresponding to the data node $v_1$. For $e_2$, because there are already two matches for $v_2$, we do not add any new data node to $G$. 

\item $[$lines 24-37$]$: For each link $e=(u,v)$ in $sm(s_i)$, we find the nodes in $G$ to which $u$ and $v$ are mapped (say $u'$ and $v'$). If $G$ includes a link with the same label as the label of $e$ between $u'$ and $v'$, we only add $s_i$ to the tags associated with the link. Otherwise, we add a new link to the graph and tag it with $s_i$.  

\end{enumerate}

\begin{figure*}[tph]
\begin{center}
\scalebox{1} {\includegraphics[width=1\linewidth]{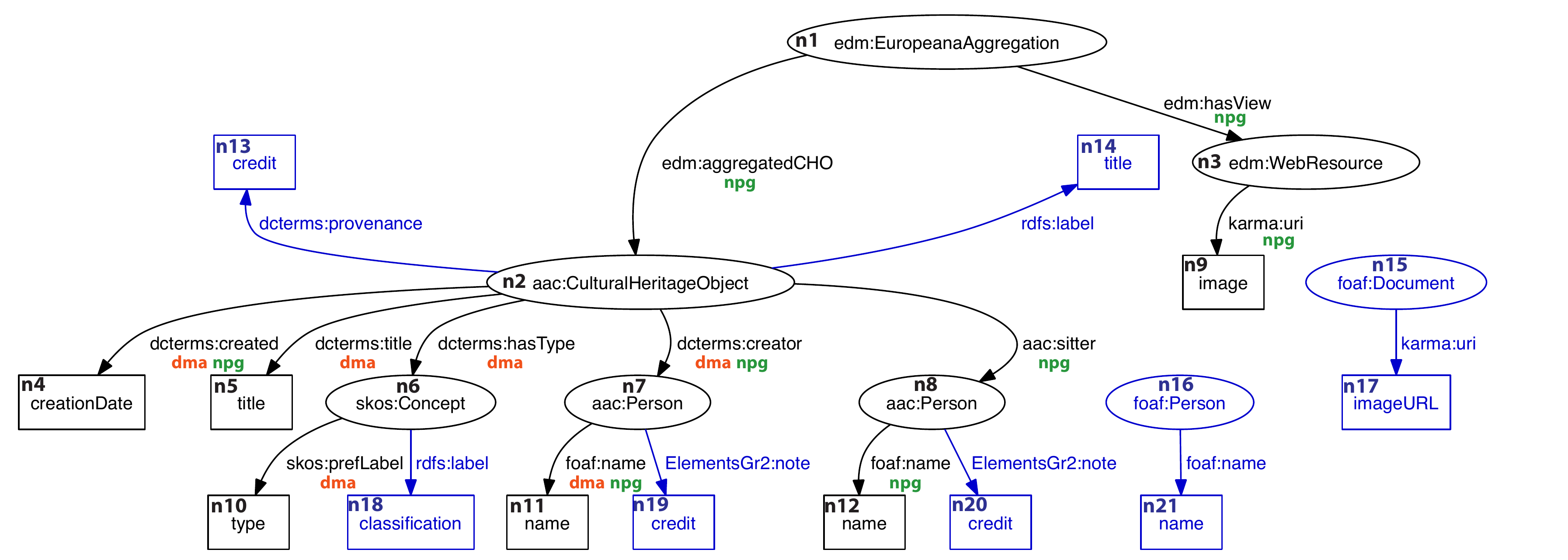}}
\caption{The graph $G$ after adding the nodes and the links corresponding to the semantic types (shown in blue).}
\label{fig:graph-step2}
\end{center}
\end{figure*}

\noindent \textbf{Adding Semantic Types}: Once the known semantic models are added to $G$, we add the semantic types learned for the attributes of the target source. As mentioned before, we have two kinds of semantic types: \textit{$\langle$class\U uri$\rangle$} for attributes whose data values are URIs and \textit{$\langle$class\U uri,property\U uri$\rangle$} for attributes that have literal data. For each learned semantic type $t$, we search the graph to see whether $G$ includes a \textit{match} for $t$. 

\begin{itemize}

	\item \textit{t=$\langle$class\U uri$\rangle$}: We say $(u,v,e)$ is a match for $t$ if $u$ is a class node with the label \textit{class\U uri}, $v$ is a data node, and $e$ is a link from $u$ to $v$ with the label \textit{karma:uri}. For example, in Figure~\ref{fig:graph-step1}, \textit{($n_{3}$,$n_{9}$,karma:uri)} is a match for the semantic type \textit{$\langle$edm:WebResource$\rangle$}.

	\item \textit{t=$\langle$class\U uri,property\U uri$\rangle$}: We say $(u,v,e)$ is a match for $t$ if $u$ is a class node labeled with \textit{class\U uri}, $v$ is a data node, and $e$ is a link from $u$ to $v$ labeled with \textit{property\U uri}. In Figure~\ref{fig:graph-step1}, \textit{($n_{6}$,$n_{10}$,skos:prefLabel)} is a match for the semantic type \textit{$\langle$skos:Concept,skos:prefLabel$\rangle$}.

\end{itemize}

We say \textit{t=$\langle$class\U uri$\rangle$} or \textit{t=$\langle$class\U uri,property\U uri$\rangle$} has a \textit{partial match} in $G$ when we cannot find a full match for $t$ but there is a class node in $G$ whose label matches \textit{class\U uri}. For instance, the semantic type \textit{$\langle$skos:Concept,rdfs:label$\rangle$} only has a partial match in $G$, because $G$ contains a class node labeled with \textit{skos:Concept} ($n_{6}$), but this class node does not have an outgoing link with the label \textit{rdfs:label}.

\begin{algorithm}[t]
	\caption{\textit{Add Semantic Types to G}}
	\label{alg:addsemantictypes}
	{\fontsize{8}{8}\selectfont
	\begin{algorithmic}[1]
	
	\Function{AddSemanticTypes}{$G$,$T$}
		\newline
    	\For  {each $a_i \in attributes(s)$} 
            \For  {each $t_{ij}^{p_{ij}} \in (t_{i1}^{p_{i1}}$, $\cdots$, $t_{ik}^{p_{ik}})$} 
            	\If {$t_{ij}=\langle class\_uri\rangle$}
            			\Let{$l_{v}$}{$class\_uri$}
            			\Let{$l_{e}$}{``$karma\colon uri$''}
            	\ElsIf {$t_{ij}=\langle class\_uri,property\_uri\rangle$}
            			\Let{$l_{v}$}{$class\_uri$}
            			\Let{$l_{e}$}{$property\_uri$}
            	\EndIf
          		\If {no node in $G$ has the label $l_{v}$}
 	      			\State add a new node $v$ with the label $l_v$ in $G$
           		\EndIf
           		\Let{$V_{match}$}{all the class nodes with the label $l_v$}
    			\Let{$w_h$}{$|E|$}
           		\For {each $v \in V_{match}$}
           			\If {$v$ does not have an outgoing link labeled $l_e$}
           				\State add a data node $w$ with the label $a_i$ to $G$
           				\State add a link $e=(v,w)$ with the label $l_e$
           				\Let{$weight(e)$}{$w_h$}
           			\EndIf
           		\EndFor
           	\EndFor
        \EndFor
    	\newline
    \EndFunction
	\end{algorithmic}}
\end{algorithm}

Algorithm~\ref{alg:addsemantictypes} shows the function that adds the learned semantic types to the graph $G$. For each semantic type $t$ learned in the labeling step, we add the necessary nodes and links to $G$ to create a match or complete existing partial matches. Consider the semantic types learned for the source \textit{dia} (Table~\ref{tab:semantic-types}). Figure~\ref{fig:graph-step2} illustrates the graph $G$ after adding the semantic types. The nodes and the links that are added in this step are depicted with the blue color. For \textit{$\langle$aac:CulturalHeritageObject, dcterms:title$\rangle$}, we do not need to change $G$, because the graph already contained one match: \textit{($n_{2}$,$n_{5}$,dcterms:title)}. The semantic type \textit{$\langle$skos:Concept,rdfs:label$\rangle$} only had one partial match ($n_{6}$), thus, we add one data node ($n_{18}$ with a label equal to the name of the corresponding attribute) and one link (\textit{rdfs:label} from $n_{6}$ to $n_{18}$) in order to complete the existing partial match. The semantic type \textit{$\langle$foaf:Document$\rangle$} had neither a match nor a partial match. We add a class node ($n_{15}$), a data node ($n_{17}$), and a link between them (\textit{karma:uri} from $n_{15}$ to $n_{17}$) to create a match. \\

\begin{figure*}[t]
\begin{center}
\scalebox{1} {\includegraphics[width=1\linewidth]{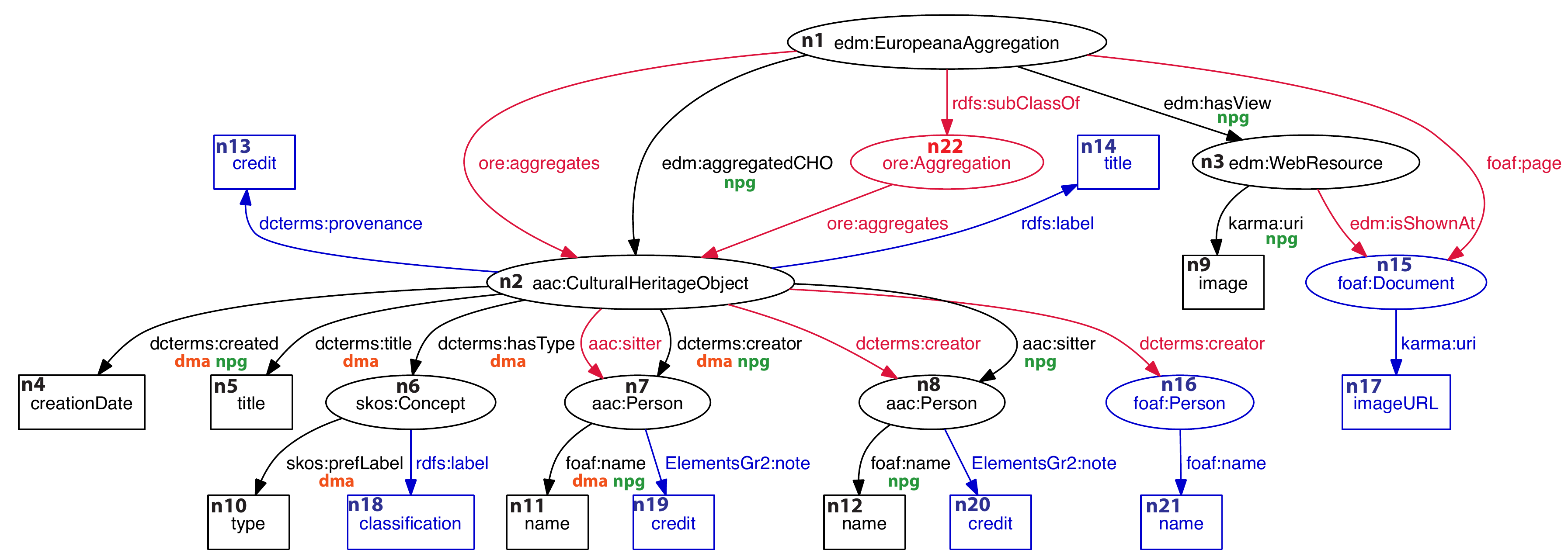}}
\caption{The final graph $G$ after adding the paths from the domain ontologies. For legibility, only a few of all the possible paths between the class nodes are shown (drawn with the red color).}
\label{fig:graph-step3}
\end{center}
\end{figure*}

\noindent \textbf{Adding Paths from the Ontology}: We use the domain ontology to find all the paths that relate the current class nodes in $G$ (Algorithm~\ref{alg:addontologypaths}). The goal is to connect class nodes of $G$ using the direct paths or the paths inferred through the subclass hierarchy in $O$. The final graph is shown in Figure~\ref{fig:graph-step3}. We connect two class nodes in the graph if there is an object property or \textit{subClassOf} relationship that connects their corresponding classes in the ontology. For instance, in Figure~\ref{fig:graph-step3}, there is the link \textit{ore:aggregates} from $n_{1}$ to $n_{2}$. This link is added because the object property \textit{ore:aggregates} is defined with \textit{ore:Aggregation} as domain and \textit{ore:AggregatedResource} as range, and \textit{edm:EuropeanaAggregation} is a subclass of the class \textit{ore:Aggregation} and \textit{aac:CulturalHeritageObject} is a subclass of \textit{edm:ProvidedCHO}, which is in turn a subclass of the class \textit{ore:AggregatedResource}. As another example, the reason why $n_{1}$ is connected to $n_{15}$ is that the property \textit{foaf:page} is defined from \textit{owl:Thing} to \textit{foaf:Document} in the FOAF ontology. Thus, a link with the label \textit{foaf:page} would exist from each class node in $G$ to $n_{15}$ since all classes are subclasses of the class \textit{owl:Thing}. Depending on the size of the ontology, many nodes and links may be added to the graph in this step. To make the figure readable, only a few of the added nodes and links are illustrated in Figure~\ref{fig:graph-step3} (the ones with the red color).

In cases where $G$ consists of disconnected components, we add a class node with the label \textit{owl:Thing} to the graph and connect the class nodes that do not have any parent to this root node using a \textit{rdfs:subClassOf} link. This converts the original graph to a graph with only one connected component.

\begin{algorithm}[t]
	\caption{\textit{Add Ontology Paths to G}}
	\label{alg:addontologypaths}
	{\fontsize{8}{8}\selectfont
	\begin{algorithmic}[1]
	
	\Function{AddOntologyPaths}{$G$,$O$}
		\newline
    	\For {each pair of class nodes $u$ and $v$ in $G$}
    		\Let{$c_1$}{ontology class with $uri=l_u$}
    		\Let{$c_2$}{ontology class with $uri=l_v$}
    		\Let{$P_{(c_1,c_2)}$}{all the direct and inferred properties (including \textit{rdfs:subClassOf}) from $c_1$ to $c_2$ in $O$}
    		\Let{$w_h$}{$|E|$}
    		\For {each property $p \in P$}
    			\Let{$l_e$}{$uri$ of the property $p$}
    			\If {there is no link with label $l_e$ from $u$ to $v$}
    				\State add a link $e=(u,v)$ with label $l_e$ to $G$
    				\Let{$weight(e)$}{$w_h$}
    			\EndIf
    		\EndFor
		\EndFor
    	\newline
    \EndFunction
	\end{algorithmic}}
\end{algorithm}

%\begin{figure*}[t]
%\centering
%\begin{subfigure}[b]{.95\textwidth}
%  \centering
%  %\imagebox{48mm}{\includegraphics[width=.95\linewidth]{figures/dma}}
%  % imagebox aligns top of the two images that are next to each other
%  \includegraphics[width=.95\linewidth]{figures/graph-step1}
%  \caption{}
%  \label{fig:graph-step1}
%\end{subfigure} \\
%% removing the new line (\\) from the end puts two images next to each other
%\begin{subfigure}[b]{.95\textwidth}
%  \centering
%  %\imagebox{48mm}{\includegraphics[width=.95\linewidth]{figures/npg}}
%  \includegraphics[width=.95\linewidth]{figures/graph-step2}
%  \caption{}
%  \label{fig:graph-step2}
%\end{subfigure} \\
%% removing the new line (\\) from the end puts two images next to each other
%\begin{subfigure}[b]{.95\textwidth}
%  \centering
%  %\imagebox{48mm}{\includegraphics[width=.95\linewidth]{figures/npg}}
%  \includegraphics[width=.95\linewidth]{figures/graph-step3}
%  \caption{}
%  \label{fig:graph-step3}
%\end{subfigure}
%\caption{}
%\label{fig:known-models}
%\end{figure*}

The links in the graph $G$ are weighted. Assigning weights to the links of the graph is very important in our algorithm. We can divide the links in $G$ into two categories. The first category includes the links that are associated with the known semantic models (black links in Figure~\ref{fig:graph-step3}). The other group consists of the links added from the learned semantic types or the ontology (blue and red links) which are not tagged with any identifier. The basis of our weighting function is to assign a much lower weight to the links in the former group compared to the links in the latter group. If $w_l$ is the default weigh of a link in the first group and $w_h$ is the default weight of a link in the second group, we will have $w_l \ll w_h$. The intuition behind this decision is to produce more coherent models in the next step when we are generating minimum-cost semantic models (Section~\ref{sec:ranking}). Our goal is to give more priority to the models containing larger segments from the known patterns. One reasonable value for $w_h$ is $w_l * |E|$ in which $|E|$ is the number of links in $G$. This formula ensures that even a long pattern from a known semantic model will cost less than a single link that does not exist in any known semantic model.

One factor that we consider in weighting the links coming from the known semantic models (black links) is the popularity of the links, i.e., the number of known semantic models supporting that link. We assign $(w_l -x/(n+1))$ to each black link where $n$ is the number of known semantic models and $x$ is the number of identifiers the link is tagged with. Suppose that we use $w_l=1$ in our example. Since our graph in Figure~\ref{fig:graph-step3} has a total of 26 links, we will have $w_h=w_l*|E|=26$. In Figure~\ref{fig:graph-step3}, the link \textit{edm:hasView} from $n_{1}$ to $n_{3}$ will be weighted with $0.66$ because it is only supported by $sm(npg)$ (\textit{n=2, x=1}). The weight of the link \textit{dcterms:creator} from $n_{2}$ to $n_{7}$ will be $0.33$ since both $sm(dma)$ and $sm(npg)$ contain that link (the link has two tags). 

We assign $w_h$ to the links that are not associated with the known models (blue and red links, which do not have a tag). There is only a small adjustment for the links coming from the ontology (red links). We prioritize direct properties over inherited properties by assigning a slightly higher weight ($w_h+\epsilon$) to the inherited ones. The rationale behind this decision comes from this observation that the direct properties (more specific) are more likely to be used in the semantic models than the inherited properties (more general). For instance, the red link \textit{aac:sitter} from $n_{2}$ to $n_{7}$ 
% $n_{6}$ to $n_{12}$ 
will be weighted with $w_h=26$, because its definition in the ontology AAC has \textit{aac:CulturalHeritageObject} as domain and \textit{aac:Person} as range. In other hand, the weight of the link \textit{ore:aggregates} from $n_{1}$ to $n_{2}$ will be $26.01$ (assume $\epsilon=0.01$) since the domain of \textit{ore:aggregates} in the ontology ORE is the class \textit{ore:Aggregation} (which is a superclass of \textit{edm:EuropeanaAggregation}) and its range is the class \textit{ore:AggregatedResource} (which is a superclass of \textit{aac:CulturalHeritageObject}).

\subsection{Mapping Source Attributes to the Graph}
\label{sec:mapping}

We use the graph built in the previous step to infer the relationships between the source attributes. First, we find mappings from the source attributes to a subset of the nodes of the graph. Then, we use these mappings to generate and rank candidate semantic models. In this section, we describe the mapping process, and in Section~\ref{sec:ranking}, we talk about computing candidate semantic models. 

To map the attributes of a source to the nodes of $G$ (Figure~\ref{fig:graph-step3}), we search $G$ to find the nodes matching the semantic types associated with the attributes. 
For example, the attribute \textit{classification} in \textit{dia} maps to $\{n_{6},n_{10}\}$ and $\{n_{6},n_{18}\}$, corresponding to the semantic types  \textit{$\langle$skos:Concept,skos:prefLabel$\rangle$} and \textit{$\langle$skos:Concept,rdfs:label$\rangle$}, respectively. 

% This attribute maps to $\{n_{6},n_{10}\}$ because \textit{($n_{6}$,$n_{10}$,skos:prefLabel)} is a match for \textit{$\langle$skos:Concept,skos:prefLabel$\rangle$} and it maps to $\{n_{6},n_{18}\}$ because \textit{($n_{6}$,$n_{18}$,rdfs:label)} is a match for \textit{$\langle$skos:Concept,rdfs:label$\rangle$}.

% For example, the attribute \textit{classification} in \textit{dia} maps to $\{n_{6},n_{10}\}$ and $\{n_{6},n_{18}\}$. Two semantic types have been assigned to the attribute \textit{classification}: \textit{$\langle$skos:Concept,skos:prefLabel$\rangle$} and \textit{$\langle$skos:Concept,rdfs:label$\rangle$}. This attribute maps to $\{n_{6},n_{10}\}$ because \textit{($n_{6}$,$n_{10}$,skos:prefLabel)} is a match for \textit{$\langle$skos:Concept,skos:prefLabel$\rangle$} and it maps to $\{n_{6},n_{18}\}$ because \textit{($n_{6}$,$n_{18}$,rdfs:label)} is a match for \textit{$\langle$skos:Concept,rdfs:label$\rangle$}.

Since each attribute has been annotated with \textit{k} semantic types and also each semantic type may have more than one match in $G$ (e.g., \textit{$\langle$aac:Person,foaf:name$\rangle$} maps to $\{n_{7},n_{11}\}$ and $\{n_{8},n_{12}\}$), more than one mapping $m$ might exist from the source attributes to the nodes of $G$. Generating all the mappings is not feasible in cases where we have a data source with many attributes and the learned semantic types have many matches in the graph. The problem becomes worse when we generate more than one candidate semantic type for each attribute. Suppose that we are modeling the source $s$ consisting of $n$ attributes and we have generated $k$ semantic types for each attribute. If there are $r$ matches for each semantic type, we will have $(k * r)^n$ mappings from the source attributes to the nodes of $G$. 
%In our example where $G$ is very small and $s$ only has five attributes, the number of possible mappings is 96.
%\yellow{szeke:} How am I supposed to interpret the example? 96 is not a large number, so what's the problem?

We present a heuristic search algorithm that explores the space of possible mappings as we map the semantic types to the nodes of the graph and expands only the most promising mappings. The algorithm scores the mappings after processing each attribute and removes the low score ones. Our scoring function takes into account the confidence values of the semantic types, the coherence of the nodes in the mappings, and the size of the mappings. The inputs to the algorithm are the learned semantic types 
$T=\{(t_{11}^{p_{11}}$, $\cdots$, $t_{1k}^{p_{1k}})$, $\cdots$, $(t_{m1}^{p_{m1}}$, $\cdots$, $t_{mk}^{p_{mk}})\}$
for the attributes of the source $s(a_1,\allowbreak a_2,\allowbreak \cdots,\allowbreak a_m)$ 
and the graph $G$, and the output is a set of candidate mappings $m$ from the source attributes to a subset of the nodes in $G$. The key idea is that instead of generating all the mappings (which is not feasible), we score the partial mappings after processing each attribute and prune the mappings with lower scores. In other words, as soon as we find the matches for the semantic types of an attribute, we rank the partial mappings and keep the better ones. In this way, the number of candidate mappings never exceeds a fixed size (\textit{branching factor}) after mapping each attribute. 

\begin{algorithm}[t]
	\caption{\textit{Generate Candidate Mappings}}
	\label{alg:mapping}
	{\fontsize{8}{8}\selectfont
	\begin{algorithmic}[1]
		\Statex \textbf{Input:} 
		\Statex ~~~~~~~~ - $G(V, E)$, 
		\Statex ~~~~~~~~ - $attributes(s)=\{a_1, \cdots, a_m\}$
		\Statex ~~~~~~~~ - $T=\{(t_{11}^{p_{11}}$, $\cdots$, $t_{1k}^{p_{1k}})$, $\cdots$, $(t_{m1}^{p_{m1}}$, $\cdots$, $t_{mk}^{p_{mk}})\}$
		\Statex ~~~~~~~~ - $branching\_factor$: max number of mappings to expand
		\Statex ~~~~~~~~ - $num\_of\_candidates$: number of candidate mappings
		\Statex \textbf{Output:} a set of candidate mappings $m$ from $attributes(s)$ to $S \subset V$
		\newline
    	\Let{$\mathit{mappings}$}{$\{\}$}
    	\Let{$\mathit{candidates}$}{$\{\}$}
    	\For  {each $a_i \in \mathit{attributes(s)}$} 
            \For  {each $t_{ij}^{p_{ij}} \in (t_{i1}^{p_{i1}}$, $\cdots$, $t_{ik}^{p_{ik}})$} 
    	        \Let{$\mathit{matches}$}{all the $(u,v,e)$ in $G$ matching $t_{ij}$}
		        \If {$\mathit{mappings=\{\}}$}
        		    \For  {each $(u,v,e) \in \mathit{matches}$} 
        				\Let{$m$}{$(\{a_i\} \rightarrow \{u, v\})$}
    	    			%\Let{$confidence(m)$}{$confidence(m) \cup p$}
    	    			\Let{$\mathit{mappings}$}{$\mathit{mappings} \cup m$}
    	    			%\State compute $score(m)$
    		    	\EndFor
			    \Else 					
     		        \For  {each $m:X \rightarrow Y \in \mathit{mappings}$}
            		    \For  {each $(u,v,e) \in \mathit{matches}$} 
            				\Let{$m'$}{($X \cup \{a_i\} \rightarrow Y \cup \{u, v\})$}
    	    			    %\Let{$confidence(m)$}{$confidence(m) \cup p$}
    	        			\Let{$\mathit{mappings}$}{$\mathit{mappings} \cup m'$}
    	    			    %\State compute $score(m')$
    		    	    \EndFor
				        \State remove $m$ from $\mathit{mappings}$
	                 \EndFor
		    	\EndIf
			\EndFor
			\If {$|\mathit{mappings}| > \mathit{branching\_factor}$}
			    \State compute $\mathit{score(m)}$ for each $m \in \mathit{mappings}$
        		\State sort items in $\mathit{mappings}$ descending based on their score
	        	\State keep top $\mathit{branching\_factor}$ mappings and remove others
	        \EndIf
		\EndFor
    	\Let{$\mathit{candidates}$}{top $\mathit{num\_of\_candidates}$ items from $\mathit{mappings}$}
		\newline
		\Return $\mathit{candidates}$
	\end{algorithmic}}
\end{algorithm}

Algorithm \ref{alg:mapping} shows our mapping process. The heart of the algorithm is the scoring function we use to rank the partial mappings (line 22 in Algorithm \ref{alg:mapping}). We compute three functions for each mapping $m$: $\mathit{confidence(m)}$, $\mathit{coherence(m)}$, and $\mathit{sizeReduction(m)}$. Then, we calculate the final score $score(m)$ by combining the values of these three functions. We explain these functions using an example. Suppose that the maximum number of the mappings we expand in each step is 2 ($\mathit{branching\_factor=2}$).
% ($branching\_factor$ in line 3). 
After mapping the second attribute of the source $dia$ ($credit$), we will have: 
$mappings=\{$ \\ 
\indent $m_1:~\{title, credit\} \rightarrow \{(n_{2},n_{5}),(n_{2},n_{13})\},$ \\
\indent $m_2:~\{title, credit\} \rightarrow \{(n_{2},n_{5}),(n_{7},n_{19})\},$ \\
\indent $m_3:~\{title, credit\} \rightarrow \{(n_{2},n_{5}),(n_{8},n_{20})\},$ \\
\indent $m_4:~\{title, credit\} \rightarrow \{(n_{2},n_{14}),(n_{2},n_{13})\},$ \\
\indent $m_5:~\{title, credit\} \rightarrow \{(n_{2},n_{14}),(n_{7},n_{19})\},$ \\
\indent $m_6:~\{title, credit\} \rightarrow \{(n_{2},n_{14}),(n_{8},n_{20})\}$ \\
 $\}$

There are two matches for the attribute $title$: $(n_{2}, n_{5})$ for the semantic type \textit{$\langle$aac:CulturalHeritageObject, dcterms:title$\rangle$} and $(n_{2}, n_{14})$ for the semantic type \textit{$\langle$aac:CulturalHeritageObject,rdfs:label$\rangle$}; and three matches for the attribute $credit$: $(n_{2}, n_{13})$ for the semantic type \textit{$\langle$aac:CulturalHeritageObject, dcterms:\-provenance$\rangle$} and $(n_{7}, n_{19})$ and $(n_{8}, n_{20})$ for the semantic type \textit{$\langle$aac:Person, ElementsGr2:note$\rangle$}. This yields $\mathit{2*3=6}$ different mappings. Since $\mathit{branching\_factor=2}$, we have to eliminate four of these mappings. Now, we describe how the algorithm ranks the mappings.  

\textbf{Confidence}: We define confidence as the arithmetic mean of the confidence values associated with a mapping. For example, $m_1$ is consisting of the matches for the semantic types \textit{$\langle$aac:CulturalHeritageObject, dcterms:title$\rangle^{0.49}$} and \textit{$\langle$aac:CulturalHeritageObject, dcterms:provenance$\rangle^{0.83}$}. Thus, $\mathit{confidence(m_1)=0.66}$.

\textbf{Coherence}: This function measures the largest number of nodes in a mapping that belong to the same known semantic model. Like the links, the nodes in $G$ are also tagged with the model identifiers although we have not shown them in Figure~\ref{fig:graph-step3}. We calculate coherence as the maximum number of the nodes in a mapping that have at least one common tag. For instance, $\mathit{coherence(m_1)=0.66}$ because two nodes out of the three nodes in $m_1$ ($n_{2}$ and $n_{5}$) are from $sm(dma)$, and $\mathit{coherence(m_2)=1.0}$ because all the nodes of $m_2$ are from the same semantic model $sm(dma)$. The goal of defining the coherence is to give more priority to the models containing larger segments from the known patterns. 

\textbf{Size Reduction}: We define the size of a mapping $\mathit{size(m)}$ as the number of the nodes in the mapping. Since we prefer concise models, we seek mappings with fewer nodes. If a mapping has $k$ attributes, the smallest possible size for this mapping is $l=k + 1$ (when all the attributes map to the same class node, e.g., $m_1$) and the largest is $u=2 * k$ (when all the attributes map to different class nodes, e.g., $m_2$). Thus, the possible size reduction in a mapping is $u-l$. We define $\mathit{sizeReduction(m)=(u-size(m))/(u-l+1)}$ as how much the size of a mapping is reduced compared to the possible size reduction. For example, $\mathit{sizeReduction(m_1)=0.5}$ and $\mathit{sizeReduction(m_2)=0}$. 

\textbf{Score(m)}: The final score is the combination the values $\mathit{confidence(m)}$, $\mathit{coherence(m)}$, and $\mathit{sizeReduction(m)}$, which are all in the range $[0,1]$. We assign a weight to each of these values and then compute the final score as the weighted sum of them: $\mathit{score(m)}= w_1~\mathit{confidence(m)} + w_2~\mathit{coherence(m)} + w_3~\mathit{sizeReduction(m)}$, where $w_1$, $w_2$, and $w_3$ are the weights, decimal values in the range $[0,1]$ summing up to 1. The proper values of the weights can be tuned by experiments. In our evaluation (Section~\ref{sec:evaluation}), we obtained better results when all the three functions contributed equally to the final score. That is, $score(m)$ is calculated as the arithmetic mean of $\mathit{confidence(m)}$, $\mathit{coherence(m)}$, and $\mathit{sizeReduction(m)}$ ($\mathit{w_1=w_2=w_3=1/3}$). 

In our example, if we use arithmetic mean to compute the final score, the scores of the 6 mappings we mentioned before are as follows: $\mathit{score(m_1) = 0.60}$, $\mathit{score(m_2) = 0.42}$, $\mathit{score(m_3) = 0.42}$, $\mathit{score(m_4) = 0.46}$, $\mathit{score(m_5) = 0.39}$, $\mathit{score(m_6) = 0.39}$. Therefore, $m_2$, $m_3$, $m_5$, and $m_6$ will be removed from the $mappings$ (line 24), and the algorithm continues to the next iteration, which is mapping the next attribute of the source \textit{dia} (\textit{classification}) to the graph. At the end, we will have maximum \textit{branching\U factor} mappings, each of them will include all the attributes. We sort these mappings based on their score and consider the top \textit{num\U of\U candidates} mappings as the candidates (Algorithm ~\ref{alg:mapping} line 27). 

%In the next part of the approach, we compute candidate semantic models for the candidate mappings. 

%$confidence(m_1)=(0.49+0.83)/2=0.66$\\
%$confidence(m_2)=(0.49+0.06)/2=0.275$\\
%$confidence(m_3)=(0.49+0.06)/2=0.275$\\
%$confidence(m_4)=(0.28+0.83)/2=0.555$\\
%$confidence(m_5)=(0.28+0.06)/2=0.17$\\
%$confidence(m_6)=(0.28+0.06)/2=0.17$\\

%
%$coherence(m_1)=0.66$\\
%$coherence(m_2)=1.0$\\
%$coherence(m_3)=1.0$\\
%$coherence(m_4)=0.33$\\
%$coherence(m_5)=1.0$\\
%$coherence(m_6)=1.0$\\

%
%$sizeReduction(m_1)=0.5$\\
%$sizeReduction(m_2)=0$\\
%$sizeReduction(m_3)=0$\\
%$sizeReduction(m_4)=0.5$\\
%$sizeReduction(m_5)=0$\\
%$sizeReduction(m_6)=0$\\

%
%$score(m_1)=(0.66+0.66+0.5)/3=0.606$\\
%$score(m_2)=(0.275+1.0+0)/3=0.425$\\
%$score(m_3)=(0.275+1.0+0)/3=0.425$\\
%$score(m_4)=(0.555+0.33+0.5)/3=0.461$\\
%$score(m_5)=(0.17+1.0+0)/3=0.39$\\
%$score(m_6)=(0.17+1.0+0)/3=0.39$\\

\subsection{Generating and Ranking Semantic Models}
\label{sec:ranking}

Once we generated candidate mappings from the source attributes to the nodes of the graph, we compute and rank candidate semantic models. To compute a semantic model for a mapping $m$, we find the minimum-cost tree in $G$ that connects the nodes of $m$. The cost of a tree is the sum of the weights on its links. This problem is known as the Steiner Tree problem \cite{winter87:steiner}. Given an edge-weighted graph and a subset of the vertices, called Steiner nodes, the goal is to find the minimum-weight tree that spans all the Steiner nodes. The general Steiner tree problem is NP-complete, however, there are several approximation algorithms \cite{winter87:steiner,takahashi1980,kou1981,mehlhorn1988} that can be used to gain a polynomial runtime complexity. 

The inputs to the algorithm are the graph $G$ and the nodes of $m$ (as Steiner nodes) and the output is a tree that we consider as a candidate semantic model for the source. For example, for the source \textit{dia} and the mapping \textit{m: \{title,credit,classification,name,imageURL\} $\rightarrow$ \{$(n_{2}$,$n_{5})$,$(n_{2}$,$n_{13})$,$(n_{6}$,$n_{10})$,$(n_{7}$,$n_{11})$,$(n_{3}$,$n_{9})$\}}, the resulting Steiner tree will be exactly as what is shown in Figure~\ref{fig:dia}, which is the correct semantic model of the source \textit{dia}. The algorithm to compute the minimal tree prefers the links that appear in the known semantic models (links with tags) because they have a much lower weight than the other links in $G$. Additionally, since the weight of a link with tags has inverse relation with its number of tags (number of known semantic models containing the link), the semantic model obtained by computing the minimal tree will contain the links that are more popular in the known semantic models.

\begin{figure}[t]
\begin{center}
\scalebox{0.9} {\includegraphics[width=.95\linewidth]{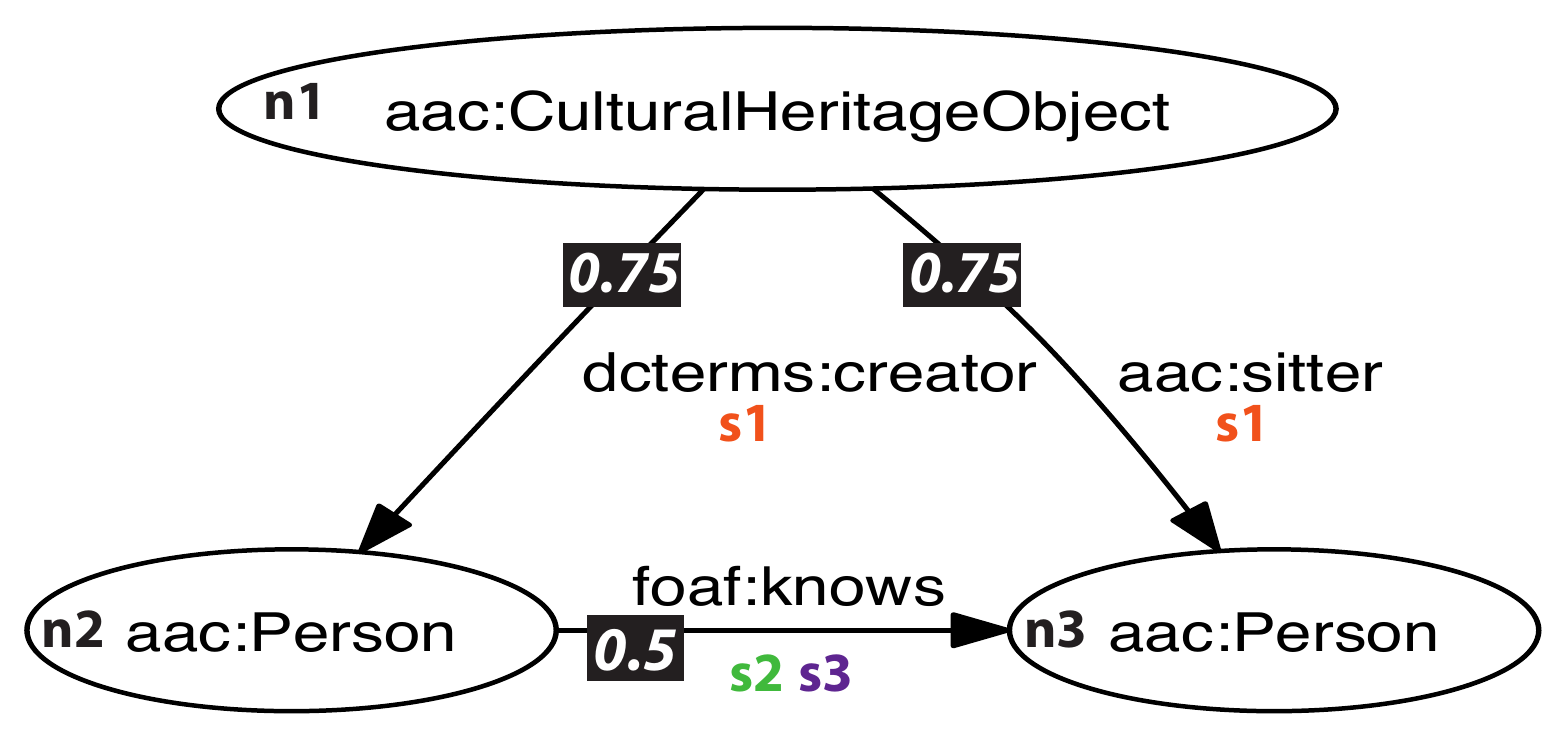}}
\caption{A small part of an example graph constructed using three known models.}
\label{fig:coherence}
\end{center}
\end{figure}

Selecting more popular links does not always yield the correct semantic model. Suppose that we have three known semantic models $\{sm(s_1),sm(s_2),sm(s_3)\}$. One of them connects \textit{aac:CulturalHeritageObject} to two instances of \textit{aac:Person} using the links \textit{dcterms:creator} and \textit{aac:sitter} (similar to $sm(npg)$). The other two semantic models do not contain the class node \textit{aac:CulturalHeritageObject}, but they have two class nodes \textit{aac:Person} connected using the link \textit{foaf:knows}. Figure~\ref{fig:coherence} shows a small part of the graph constructed using these known models. The black labels on the links represent the weights of the links. For instance, the link \textit{dcterms:creator} from $n_1$ to $n_2$ has a weight equal to $0.75$ because it is only supported by $sm(s_1)$ ($w_l-x/(n+1)=1-1/(1+3)=0.75$).  

Now, assume that we have a new source $s_4$ with three attributes $\{a_1,a_2,a_3\}$ annotated with \textit{aac:CulturalHeritageObject}, \textit{aac:Person}, and \textit{aac: Person}. Computing the minimal tree for the mapping \textit{m: $\{a_1,a_2,a_3\}$ $\rightarrow$ $\{n_1,n_2,n_3\}$} will result a tree that consists of the link \textit{foaf:knows} between $n_2$ to $n_3$ and either \textit{dcterms:creator} from $n_1$ to $n_2$ or \textit{aac:sitter} between $n_1$ and $n_3$. Nonetheless, this is not the correct semantic model for the source. When $s_4$ includes \textit{aac:CulturalHeritageObject} in addition to those two \textit{aac:Person}, it is more likely that the source is describing the relations between the cultural heritage objects and the people and not the relations between the people. 

We solve this problem by taking into account the coherence of the patterns. Instead of just the minimal Steiner tree, we compute the \textit{top-k} Steiner trees and rank them first based on the coherence of their links and then their cost. In the example shown in Figure~\ref{fig:coherence}, the top-3 results assuming $n_1$, $n_2$, and $n_3$ as the Steiner nodes are:

\indent \textit{$T_1$=\{($n_1$,$n_3$,aac:sitter),($n_2$,$n_3$,foaf:knows)\}} \\
\indent \textit{$T_2$=\{($n_1$,$n_2$,dcterms:creator),($n_2$,$n_3$,foaf:knows)\}} \\
\indent \textit{$T_3$=\{($n_1$,$n_2$,dcterms:creator),($n_1$,$n_3$,aac:sitter)\}}

\noindent where \textit{cost(T\SB 1)=cost(T\SB 2)=1.25} and \textit{cost(T\SB 3)=1.5}. Once we computed the \textit{top-k} trees, we sort them according to their coherence. The coherence here means the percentage of the links in the Steiner tree that are supported by the same semantic model. It is computed similar to the coherence of the nodes with the difference that we use the tags on the links instead of the tags on the nodes. In our example, the coherence of $T_1$ and $T_2$ will be $0.5$ because their links do not belong to the same known semantic model, and the coherence of $T_3$ will be $1.0$ since both of its links are tagged with $s_1$. Therefore, $T_3$ will be ranked higher than $T_1$ and $T_2$, although it has higher cost than $T_1$ and $T_2$. 

We use a customized version of the BANKS algorithms \cite{bhalotia2002} to compute the \textit{top-k} Steiner trees. The original BANKS algorithm is developed for the problem of the keyword-based search in relational databases, and because it makes specific assumptions about the topology of the graph, applying it directly to our problem eliminates some of the trees from the results. For instance, if two nodes are connected using two links with different weights, it only considers the one with the lower weight and it never generates a tree including the link with the higher weight. We customized the original algorithm to support more general cases.

The BANKS algorithm creates one iterator for each of the nodes corresponding to the semantic types, and then the iterators follow the incoming links to reach a common ancestor. The algorithm uses the iterator's distance to its starting point to decide which link should be followed next. Because our weights have an inverse relation with their popularity, the algorithm prefers more frequent links. To make the algorithm converge to more coherent models first, we use a heuristic that prefers the links that are parts of the same pattern (known semantic model) even if they have higher weights. Suppose that $sm_1: v_1 \overset{e_1}{\longrightarrow} v_2$ and $sm_2: v_1 \overset{e_2}{\longrightarrow} v_2 \overset{e_3}{\longrightarrow} v_3$ are the only known models used to build the graph $G$, and the weight of the link $e_2$ is higher than $e_1$. Assume that $v_1$ and $v_3$ are the semantic labels. The algorithm creates two iterators, one starting from $v_1$ and one from $v_3$. The iterator that starts from $v_3$ reaches $v_2$ by following the incoming link $v_2 \overset{e_3}{\longrightarrow} v_3$. At this point, it analyzes the incoming links of $v_2$ and although $e_1$ has lower weight than $e_2$, it first chooses $e_2$ to traverse next. This is because $e_2$ is part of the known model $sm_2$ which includes the previously traversed link $e_3$.

It is important to note that considering coherence of patterns in scoring the mappings and also ranking the final semantic models enables our approach to compute the correct semantic model in many cases where the top semantic types are not the correct ones. For example, for the source \textit{dia}, the mapping \textit{m:\{title,credit,classification,name,imageURL\} $\rightarrow$ \{$(n_{2}$,$n_{5})$,$(n_{2}$,$n_{13})$,$(n_{6}$,$n_{10})$,$(n_{7}$,$n_{11})$,$(n_{3}$,$n_{9})$\}},~which~maps the attribute \textit{imageURL} to $(n_{3},n_{9})$ using~the~type \textit{$\langle$edm:WebResource$\rangle$}, will~be~scored~higher~than the mapping \textit{$m'$:\{title,credit,classification,name,imageURL \}$\rightarrow$\{$(n_{2}$,$n_{5})$,$(n_{2}$,$n_{13})$,$(n_{6}$,$n_{10})$,$(n_{7}$,$n_{11})$,$(n_{15}$,$n_{17})$\}},~which maps \textit{imageURL} to $(n_{15},n_{17})$ using the type \textit{$\langle$foaf:Document$\rangle$}. The mapping $m$ has lower confidence value than $m'$, but is scored higher because its coherence value is higher. The model computed from the mapping $m$ will also be ranked higher than the model computed from $m'$, because it includes more links from known patterns, thus~resulting~in~a~lower~cost~tree.
\section{Evaluation}
\label{sec:evaluation}

We evaluated our approach on two datasets, each including a set of data sources and a set of domain ontologies that will be used to model the sources. Both of these datasets have the same set of data sources, 29 museum sources in CSV, XML, or JSON format containing data from different art museums in the US, however, they include different domain ontologies. The goal is to learn the semantic models of the data sources with respect to two well-known data models in the museum domain: Europeana Data Model (EDM),\footnote{\url{http://pro.europeana.eu/page/edm-documentation}} and CIDOC Conceptual Reference Model (CIDOC-CRM).\footnote{\url{http://www.cidoc-crm.org}} These data models use different domain ontologies to represent knowledge in the museum domain. 

The first dataset, $\mathit{ds}_{edm}$, contains the EDM, AAC, SKOS, Dublin Core Metadata Terms, FRBR, FOAF, ORE, and ElementsGr2 ontologies, and the second dataset, $\mathit{ds}_{crm}$, includes the CIDOC-CRM and SKOS ontologies. The reason why we used two data models is to evaluate how our approach performs with respect to different representations of knowledge in a domain. We applied our approach on both datasets to find the candidate semantic models for each source and then compared the best suggested models (the first ranked models) with models created manually by domain experts. Table~\ref{tab:datasets} shows more details of the evaluation datasets. The datasets including the sources, the domain ontologies, and the gold standard models are available on GitHub.\footnote{\url{https://github.com/taheriyan/jws-knowledge-graphs-2015}} The source code of our approach is integrated into Karma which is available as open source.\footnote{\url{https://github.com/usc-isi-i2/Web-Karma}}

Manually constructing semantic models, in addition to being time-consuming and error-prone, requires a thorough understanding of the domain ontologies. Karma \cite{knoblock12:semi} provides a user friendly graphical interface enabling users to interactively build the semantic models. Yet, building the models in Karma without any automation requires significant user effort. Our automatic approach learns accurate semantic models that can be transformed to the gold standard models by only a few user actions.

\begin{table}[t]
\caption{The evaluation datasets $\mathit{ds}_{edm}$ and $\mathit{ds}_{crm}$.}
%\vspace{-0.1in}
\begin{center}
\scalebox{0.9} {
\begin{tabular}{ | l | c | c | }
\hline
 & \textbf{$\mathit{ds}_{edm}$} & \textbf{$\mathit{ds}_{crm}$} \\ \hline
\#data source  & 29 & 29 \\ \hline
\#classes in the domain ontologies  & 119 & 147 \\ \hline
\#properties in the domain ontologies  & 351 & 409 \\ \hline
\#nodes in the gold-standard models  & 473 & 812 \\ \hline
\#data nodes in the gold-standard models  & 331 & 418 \\ \hline
\#class nodes in the gold-standard models  & 142 & 394 \\ \hline
\#links in the gold-standard models  & 444 & 785 \\ \hline
%#links between class nodes in the gold-standard models  & 113 \\ \hline
\end{tabular}
}
\end{center}
\label{tab:datasets}
\end{table}

In each dataset, we applied our method to learn a semantic model for a target source $s_i$, $sm(s_i)$, assuming that the semantic models of the other sources are known. To investigate how the number of the known models influences the results, we used variable number of known models as input. Suppose that M\SB{j} is a set of known semantic models including $j$ models. Running the experiment with M\SB{0} means that we do not use any knowledge other than the domain ontology and running it with M\SB{28} means that the semantic models of all the other sources are known (M\SB{28} is leave-one-out cross validation). For example, for $s_1$, we ran the code 29 times using M\SB{0}=$\{\}$, M\SB{1}=$\{sm(s_2)\}$, M\SB{2}=$\{sm(s_2),\allowbreak sm(s_3)\}$, $\cdots$, M\SB{28}=$\{sm(s_2),\allowbreak \cdots,\allowbreak sm(s_{29})\}$. 

In learning the semantic types of a source $s_i$, we use the data of the sources whose semantic models are known as training data. More precisely, when we are running our labeling algorithm on source $s_i$ with M\SB{j} setting, the training data is the data of all the sources $\{s_k|k=1,..,j$ \textit{and} $k!=i\}$ and the test data is the data of the target source $s_i$. Using M\SB{0} means that there is no training data and thus the labeling function will not be able to suggest any semantic type for the source attributes. To evaluate the labeling algorithm, we use \textit{mean reciprocal rank} (MRR) \cite{craswell09}, which is useful when we consider top $k$ semantic types. MRR helps to analyze the ranking of predictions made by any semantic labeling approach using a single measure rather than having to analyze top-1 to top-k prediction accuracies separately, which is a cumbersome task. In learning the semantic types of a source $s_i$ with $n$ attributes, MRR is computed as: $$MRR=\frac{1}{n} \sum_{i=1}^{n}\frac{1}{rank_i}$$ where $rank_i$ is the rank of the correct semantic type in the top $k$ predictions made for the attribute $a_i$. It is obvious that if we only consider the top semantic type predictions, the value of MRR is equal to the accuracy. In our example in Table~\ref{tab:semantic-types}, $MRR=1/5 (1/1 + 1/1 + 1/1 + 1/1 + 1/2)=0.9$ (the correct semantic type for the attribute $imageURL$ is ranked second).

We compute the accuracy of the learned semantic models by comparing them with the gold standard models in terms of \textit{precision} and \textit{recall}. Assuming that the correct semantic model of the source $s$ is $sm$ and the semantic model learned by our approach is $sm'$, we define precision and recall as:

%\vspace{-0.1in} 
%\[\Scale[1]{precision=\frac{rel(sm) \cap rel(sm')}{rel(sm')},~~ recall=\frac{rel(sm) \cap rel(sm')}{rel(sm)}}\]

$$ precision=\frac{|rel(sm) \cap rel(sm')|}{|rel(sm')|}$$
$$ recall=\frac{|rel(sm) \cap rel(sm')|}{|rel(sm)|} $$

\noindent where $\mathit{rel(sm)}$ is the set of triples $(u,v,e)$ in which $e$ is a link from the node $u$ to the node $v$ in the semantic model $sm$. For example, for the semantic model in Figure~\ref{fig:dia}, %
%$P_1=\{(Europeana\-Aggregation$ $\xrightarrow{aggregatedCHO}$ $ CulturalHeritageObject)$,
%$(Europeana\-Aggregation$ $\xrightarrow{hasView}$ $WebResource)$,
%$(CulturalHeritageObject$ $\xrightarrow{creator}$ $Person)$,$\cdots\}$. 
%
%$P_1=\{(edm \colon Europeana\-Aggregation$ $\xrightarrow{edm \colon aggregatedCHO}$ $aac \colon CulturalHeritageObject)$,
%$(edm \colon Europeana\-Aggregation$ $\xrightarrow{edm \colon hasView}$ $edm \colon WebResource)$,
%$(aac \colon CulturalHeritageObject$ $\xrightarrow{dcterms \colon creator}$ $aac \colon Person)$,$\cdots\}$. 
%
\textit{rel(sm)=\{(edm:EuropeanaAggregation, aac:\linebreak CulturalHeritageObject, edm:aggregatedCHO), 
(edm:\linebreak EuropeanaAggregation, edm:WebResource, edm:\linebreak hasView),(aac:CulturalHeritageObject, aac:Person, dcterms:creator), $\cdots$\}}. 
 
\begin{figure}[t]
\begin{center}
\scalebox{1} {\includegraphics[width=1\linewidth]{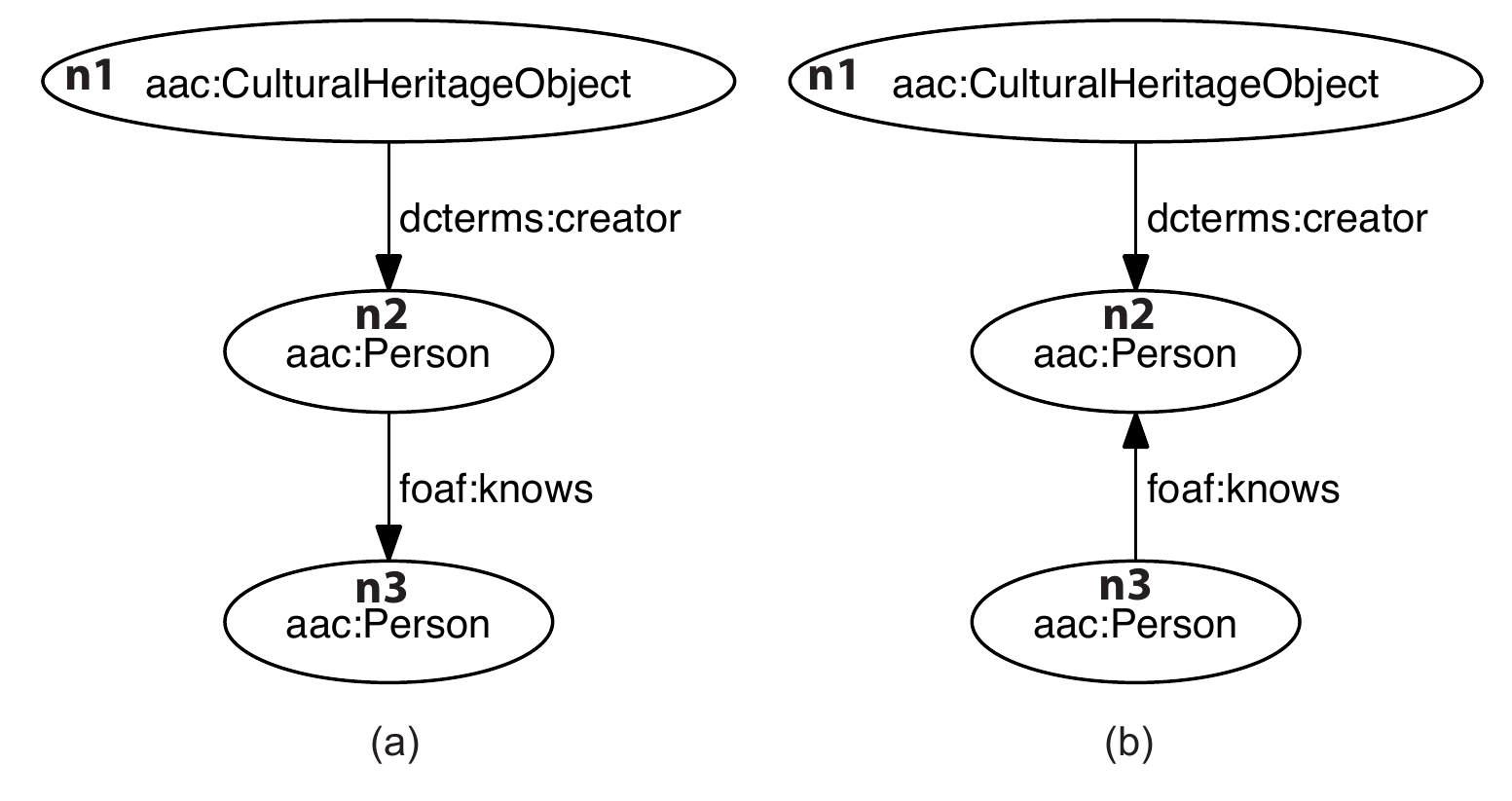}}
\caption{These two semantic models are not equivalent.}
\label{fig:evaluation-sample}
\end{center}
\end{figure}

If all the nodes in $sm$ have unique labels and all the nodes in $sm'$ also have unique labels, $\mathit{rel(sm)=rel(sm')}$ ensures that $sm$ and $sm'$ are equivalent. However, if the semantic models have more than one instance of an ontology class, we will have nodes with the same label. In this case, $\mathit{rel(sm)=rel(sm')}$ does not guarantee $\mathit{sm=sm'}$. For example, the two semantic models exemplified in Figure~\ref{fig:evaluation-sample} have the same set of triples although they do not convey the same semantics. In Figure~\ref{fig:evaluation-sample}a, the creator of the artwork knows another person while the semantic model in Figure~\ref{fig:evaluation-sample}b states that the creator of the artwork is known by another person. Many sources in our datasets have models that include two or more instances of an ontology class.

%\begin{figure}[tph]
%\centering
%\begin{subfigure}[b]{.5\textwidth}
%  \centering
%  %\imagebox{48mm}{\includegraphics[width=.95\linewidth]{figures/dma}}
%  % imagebox aligns top of the two images that are next to each other
%  \includegraphics[width=.95\linewidth]{figures/evaluation-sample1}
%  \caption{}
%  \label{fig:evaluation-sample1}
%\end{subfigure} \\
%% removing the new line (\\) from the end puts two images next to each other
%\begin{subfigure}[b]{.5\textwidth}
%  \centering
%  %\imagebox{48mm}{\includegraphics[width=.95\linewidth]{figures/npg}}
%  \includegraphics[width=.95\linewidth]{figures/evaluation-sample2}
%  \caption{}
%  \label{fig:evaluation-sample2}
%\end{subfigure}
%\caption{These two semantic models are not equivalent.}
%\label{fig:evaluation-sample}
%\end{figure}

To have a more accurate evaluation, we number the nodes and then use the numbered labels in measuring the precision and recall. Assume that the model in Figure~\ref{fig:evaluation-sample}a is the correct semantic model ($sm$) and the one in \ref{fig:evaluation-sample}b is the model learned by our approach ($sm'$). We change the labels of the nodes $n_1$, $n_2$ and $n_3$ in $sm$ to \textit{aac:CulturalHeritageObject1}, \textit{aac:Person1} and \textit{aac:Person2}. After this change, we will have \textit{rel(sm)=\{(aac:CulturalHeritageObject1, aac:Person1, dcterms:creator), 
(aac:Person1, aac:Person2, foaf: knows)\}}. Then, we try all the permutations of the numbering in the learned model $sm'$ and report the precision and recall of the one that generates the best \textit{F1-measure}.\footnote{\textit{F1-measure=}$2*(precision\times recall)/(precision+recall)$} For instance, if we number the nodes $n_2$ and $n_3$ in $sm'$ with \textit{aac:Person1} and \textit{aac:Person2}, we will have \textit{rel($sm'$)=\{(aac:CulturalHeritageObject1, aac: Person1, dcterms:creator), 
(aac:Person2, aac:Person1, foaf:knows)\}}, which yields \textit{precision=recall=0.5}. If we label $n_2$ with \textit{aac:Person2} and $n_3$ with \textit{aac:Person1}, we will have \textit{rel($sm'$)=\{(aac:CulturalHeritageObject1, aac:Person2, dcterms:creator), 
(aac:Person1, aac:Person2, foaf:knows)\}}, which still has \textit{precision=recall=0.5}. 

One of the factors influencing the results of our method is the overlap between the known semantic models and the semantic model of the target source. To see how much two semantic models overlap each other, we define the \textit{overlap} metric as the Jaccard similarity between their relationships:

$$ overlap=\frac{|rel(sm) \cap rel(sm')|}{|rel(sm) \cup rel(sm')|}$$

Table~\ref{tab:overlap} reports the minimum, maximum, median, and average overlap between the semantic models of each dataset. Overall, the higher the overlap between the known semantic models and the semantic model of the target source, the more accurate models can be learned.
\begin{table}[t]
\caption{The overlap between the pairs of the semantic models in the datasets $\mathit{ds}_{edm}$ and $\mathit{ds}_{crm}$.}
%\vspace{-0.1in}
\begin{center}
\scalebox{0.9} {
\begin{tabular}{ | l | c | c | }
\hline
 & \textbf{$\mathit{ds}_{edm}$} & \textbf{$\mathit{ds}_{crm}$} \\ \hline
minimum overlap & 0.04 & 0.03 \\ \hline
maximum overlap & 1 & 1 \\ \hline
median overlap & 0.45 & 0.46 \\ \hline
average overlap & 0.43 & 0.46 \\ \hline
\end{tabular}
}
\end{center}
\label{tab:overlap}
\end{table}

In our mapping algorithm (Algorithm~\ref{alg:mapping}), we used 50 as cut-off (\textit{branching\U factor=50}) and then considered all the generated mappings as the candidate mappings (\textit{num\U of\U candidates=50}). We justify this choice in Section~\ref{subsec:scenario2} by analyzing the impact of the branching factor on the accuracy of the results and the running time of the algorithm. To score a mapping $m$, we assigned equal weights to the functions $\mathit{confidence(m)}$, $\mathit{coherence(m)}$, and $\mathit{sizeReduction(m)}$. We tried different combinations of weights, and although our algorithm generated more precise models for a few sources in some of these weight systems, the average results were better when each of these functions contributed equally to the final score. Once we found the candidate mappings, we generated the top 10 Steiner trees for each of them (\textit{k=10} in \textit{top-k} Steiner tree algorithm). Finally, we ranked the candidate semantic models (at most 500) and compared the best one with the correct model of the source. We ran two experiments with different scenarios that will be explained next.

\subsection{Scenario 1}
\label{subsec:scenario1}

\begin{figure}
\centering
\begin{subfigure}[b]{.48\textwidth}
  \centering
  \includegraphics[width=1\linewidth]{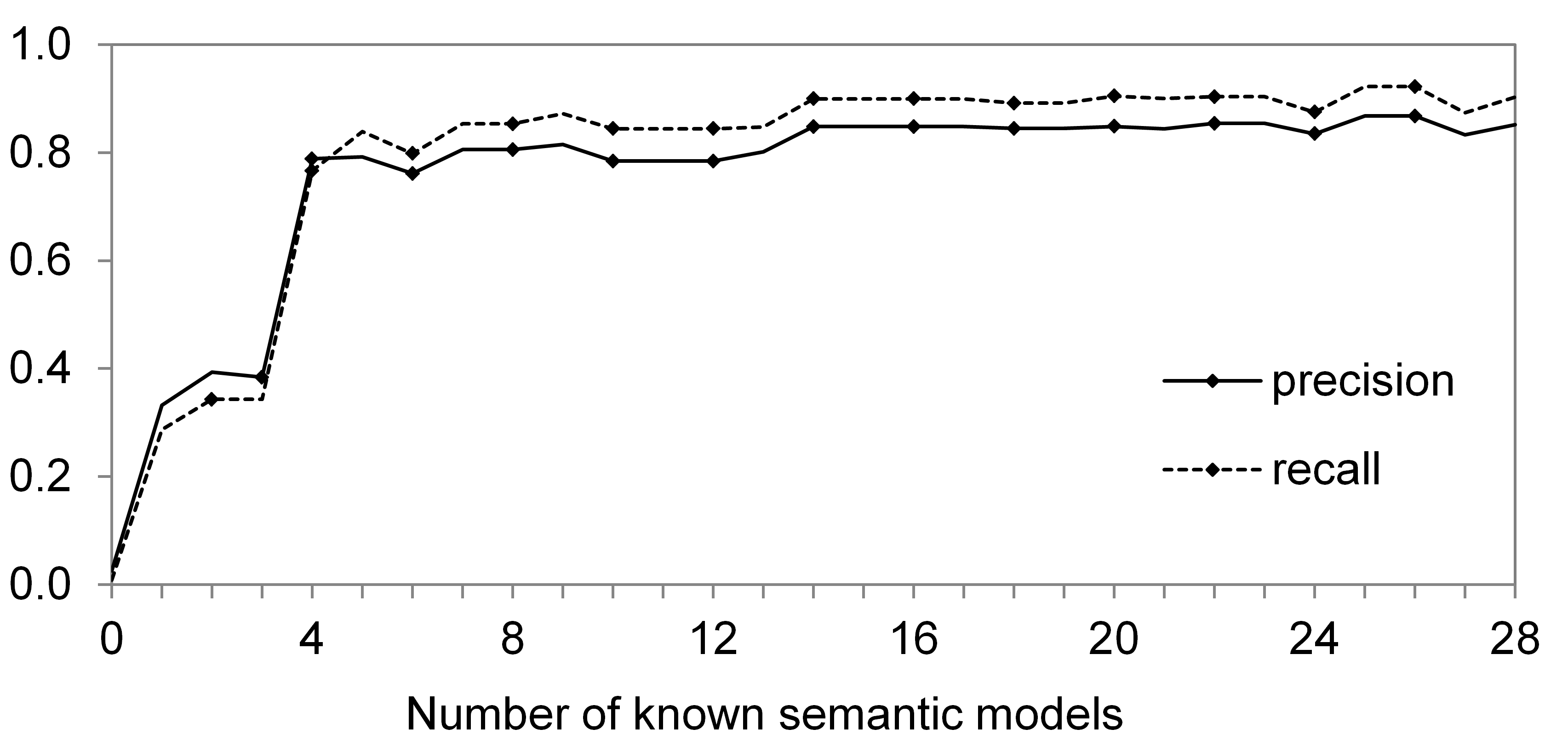}
  \caption{$\mathit{ds}_{edm}$}
  \label{fig:scenario1-edm-accuracy}
\end{subfigure} \\
% removing the new line (\\) from the end puts two images next to each other
\begin{subfigure}[b]{.48\textwidth}
  \centering
  \includegraphics[width=1\linewidth]{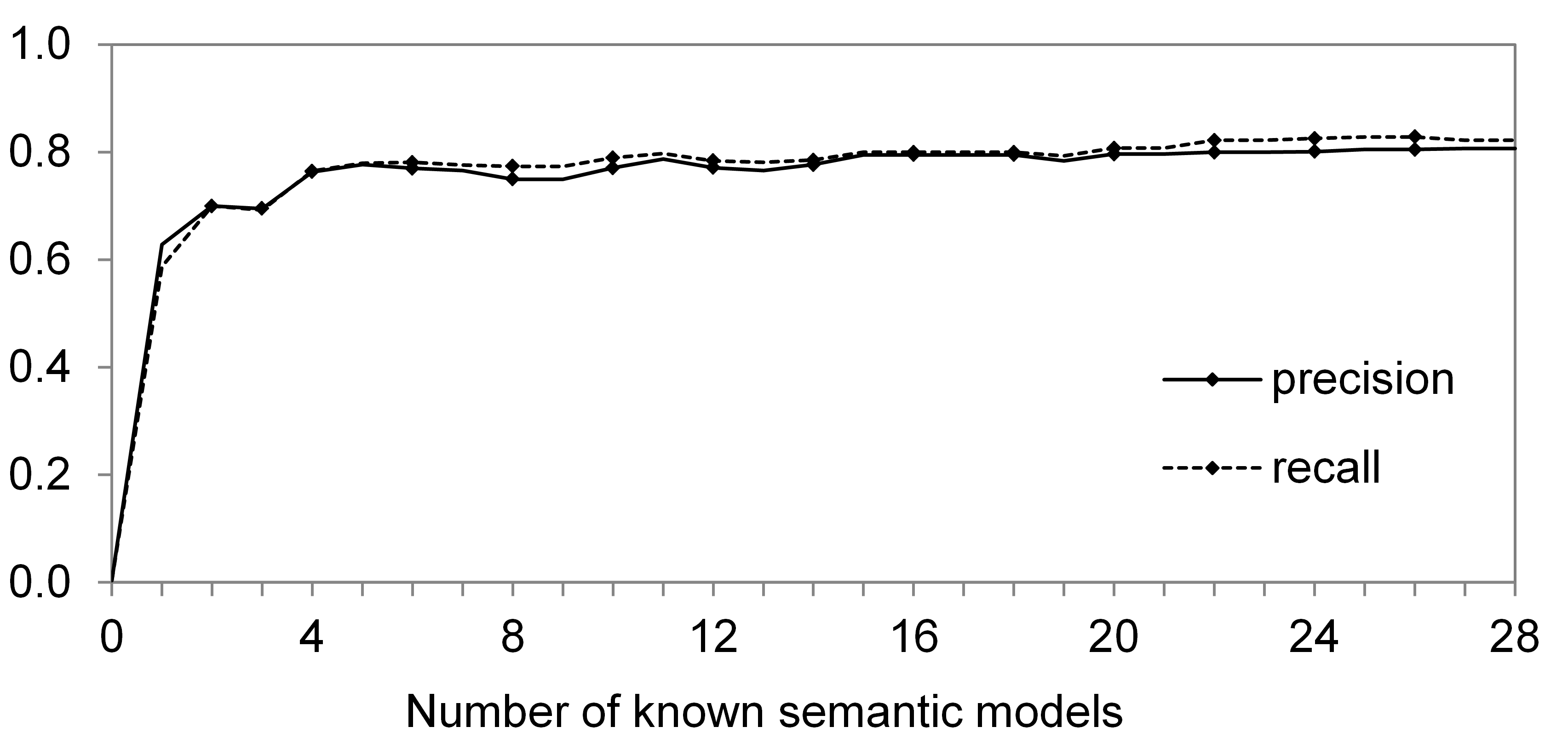}
  \caption{$\mathit{ds}_{crm}$}
  \label{fig:scenario1-crm-accuracy}
\end{subfigure}
\caption{Average precision and recall for the learned semantic models when the attributes are labeled with their correct semantic types.}
\vspace{0.05in}
\label{fig:scenario1-accuracy}
\end{figure}

In the first scenario, we assumed that each source attribute is annotated with its correct semantic type. The goal was to see how well our approach learns the attribute relationships using the correct semantic types. Figure~\ref{fig:scenario1-accuracy} illustrates the average precision and recall of all the learned semantic models ($sm'(s_1),\allowbreak \cdots,\allowbreak sm'(s_{29})$) for each M\SB{j} ($j \in [0..28]$) for each dataset. Since the correct semantic types are given, we excluded their corresponding triples in computing the precision and recall. That is, we compared only the links between the class nodes in the gold standard models with the links between the class nodes in the learned models. We call such links \textit{internal links}, the links that are established between the class nodes in semantic models. The total number of the links in the dataset $\mathit{ds}_{edm}$ is 444, and 331 of these links corresponds to the source attributes (there are 331 data nodes). Thus, $\mathit{ds}_{edm}$ has 113 internal links (444-331=113). Following the same rationale, $\mathit{ds}_{crm}$ has 367 internal links.

The results show that the precision and recall increase significantly even with a few known semantic models. An interesting observation is that when there is no known semantic model and the only background knowledge is the domain ontology (baseline, M\SB{0}), the precision and recall are close to 0. This low accuracy comes from the fact that there are multiple links between each pair of class nodes in the graph $G$, and without additional information, we cannot resolve the ambiguity. Although we assign lower weights to direct properties to prioritize them over inherited ones, it cannot help much because for many of the class nodes in the correct models, there is no object property in the ontology that is explicitly defined with the corresponding classes as domain and range. In fact, most of the properties that have been used in the correct models are either inherited properties or defined without a domain or/and range in the ontology.   %\\

%\begin{figure}
%\begin{center}
%\scalebox{0.9} {\includegraphics[width=.95\linewidth]{figures/scenario1-edm-accuracy}}
%\caption{Average precision and recall for the learned semantic models for $ds_{edm}$ when the attributes are labeled with their correct semantic types}
%\label{fig:scenario1-edm-accuracy}
%\end{center}
%\end{figure}
%
%\begin{figure}
%\begin{center}
%\scalebox{0.9} {\includegraphics[width=.95\linewidth]{figures/scenario1-crm-accuracy}}
%\caption{Average precision and recall for the learned semantic models for $ds_{crm}$ when the attributes are labeled with their correct semantic types}
%\label{fig:scenario1-crm-accuracy}
%\end{center}
%\end{figure}

\begin{figure}
\centering
\begin{subfigure}[b]{.48\textwidth}
  \centering
  \includegraphics[width=1\linewidth]{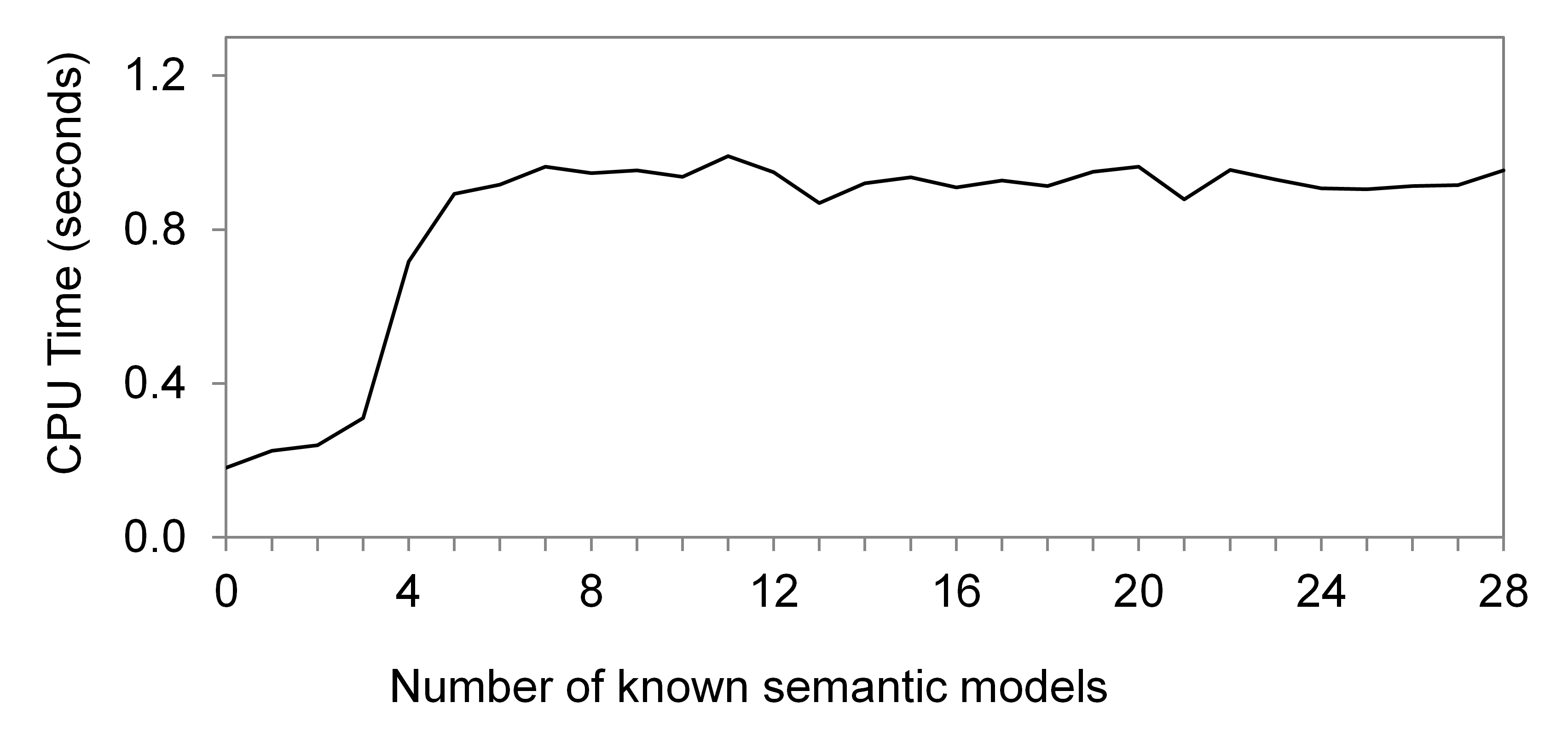}
  \caption{$\mathit{ds}_{edm}$}
  \label{fig:scenario1-edm-time}
\end{subfigure} \\
% removing the new line (\\) from the end puts two images next to each other
\begin{subfigure}[b]{.48\textwidth}
  \centering
  \includegraphics[width=1\linewidth]{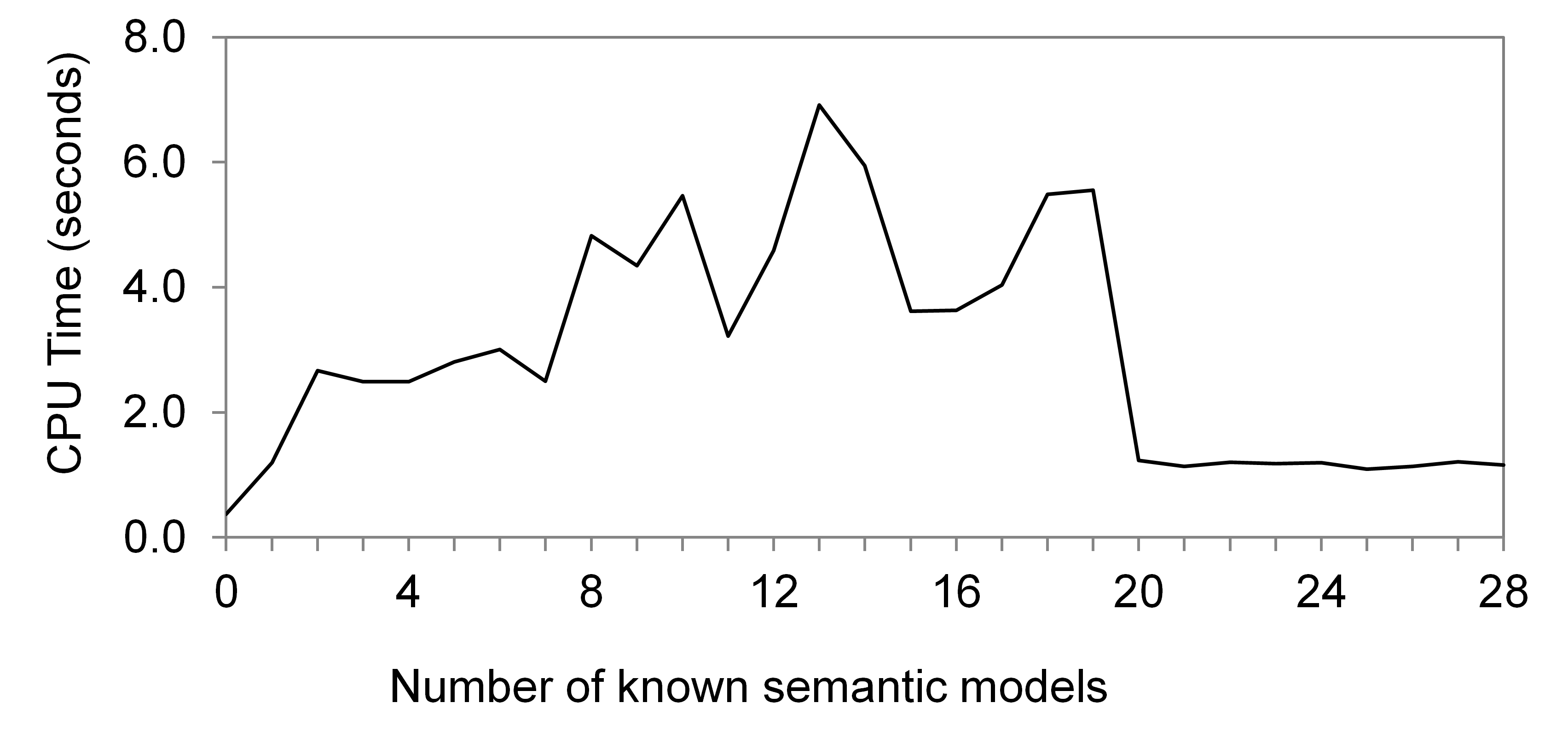}
  \caption{$\mathit{ds}_{crm}$}
  \label{fig:scenario1-crm-time}
\end{subfigure}
\caption{Average semantic model learning time when the attributes are labeled with their correct semantic types.}
\vspace{0.05in}
\label{fig:scenario1-time}
\end{figure}

To evaluate the running time of the approach, we measured the running time of the algorithm starting from building the graph until ranking the results on a single machine with a Mac OS X operating system and a 2.3 GHz Intel Core i7 CPU. Figure~\ref{fig:scenario1-time} shows the average time (in seconds) of learning the semantic models. The reason why there is some fluctuations in the timing diagram of $\mathit{ds}_{crm}$ (Figure~\ref{fig:scenario1-crm-time}) is related to the topology of the graph built on top of the known models and also the details of our implementation. While one expects to see linear increase in time when the number of known semantic models grows, sometimes adding a new semantic model changes the structure of the graph in a way that the Steiner tree algorithm finds $k$ candidate trees faster.

We believe that the overall time of the process can be further reduced by using parallel programming and some optimizations in the implementation. For example, the graph can be built incrementally. When a new known model is added, we do not need to create the graph from scratch. We just need to merge the new known model to the existing graph and update the links. 

%\begin{figure}
%\begin{center}
%\scalebox{1} {\includegraphics[width=.95\linewidth]{figures/scenario1-edm-time}}
%\caption{Average semantic model learning time for $ds_{edm}$ when the attributes are labeled with their correct semantic types}
%\label{fig:scenario1-edm-time}
%\end{center}
%\end{figure}
%
%\begin{figure}
%\begin{center}
%\scalebox{1} {\includegraphics[width=.95\linewidth]{figures/scenario1-crm-time}}
%\caption{Average semantic model learning time for $ds_{crm}$ when the attributes are labeled with their correct semantic types}
%\label{fig:scenario1-crm-time}
%\end{center}
%\end{figure}

\subsection{Scenario 2} 
\label{subsec:scenario2}

\begin{figure}
\centering
\begin{subfigure}[b]{.48\textwidth}
  \centering
  \includegraphics[width=1\linewidth]{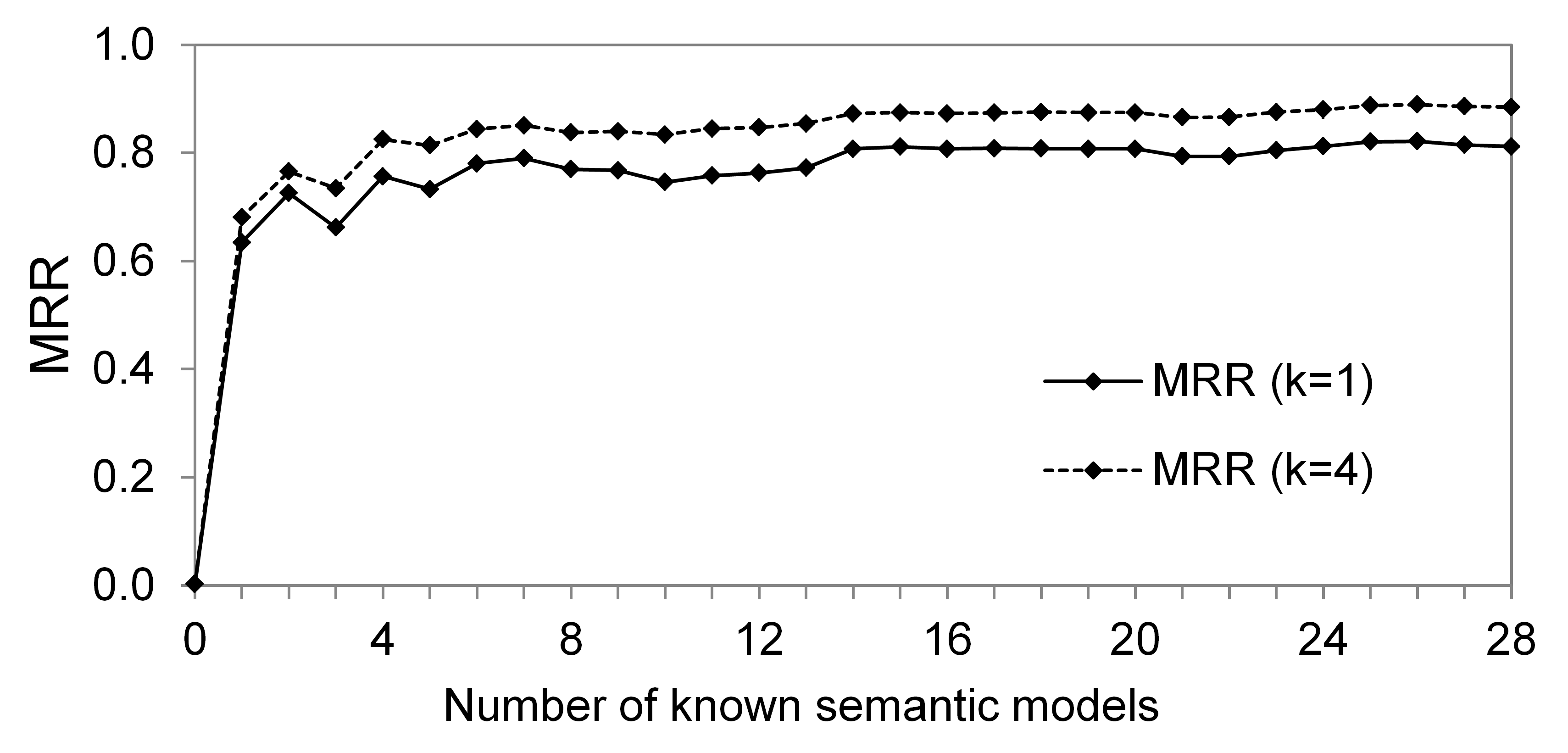}
  \caption{$\mathit{ds}_{edm}$}
  \label{fig:scenario2-edm-mrr}
\end{subfigure} \\
% removing the new line (\\) from the end puts two images next to each other
\begin{subfigure}[b]{.48\textwidth}
  \centering
  \includegraphics[width=1\linewidth]{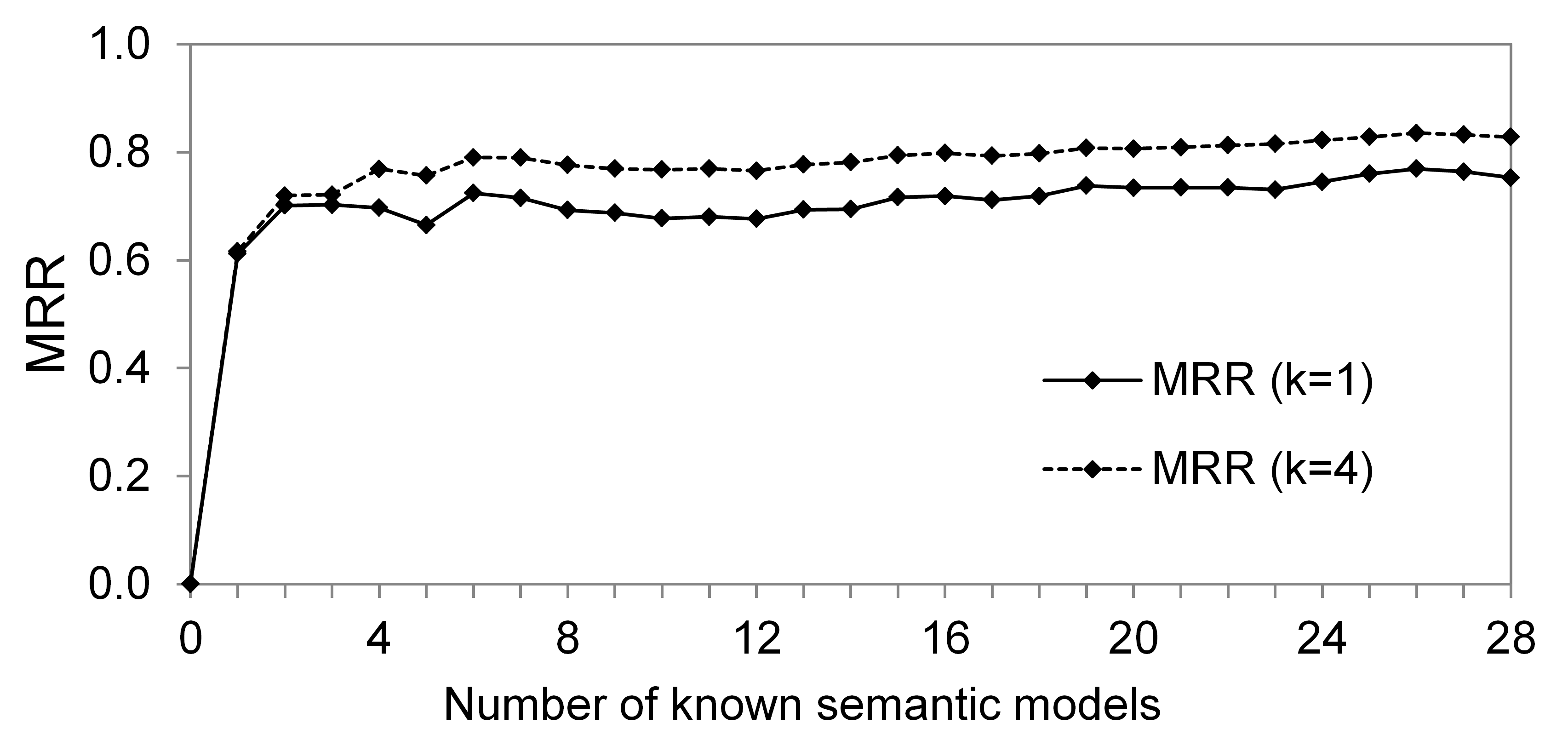}
  \caption{$\mathit{ds}_{crm}$}
  \label{fig:scenario2-crm-mrr}
\end{subfigure}
\caption{MRR value of the learned semantic types when only the top learned semantic types are considered (k=1); and the top four suggested types are considered (k=4).}
\vspace{0.05in}
\label{fig:scenario2-mrr}
\end{figure}

In the second scenario, we used our semantic labeling algorithm to learn the semantic types. We trained the labeling classifier on the data of the sources whose semantic models are already known and then applied the learned labeling function to the target source to assign a set of candidate semantic types to each source attribute. Figure~\ref{fig:scenario2-mrr} shows the MRR diagram for $\mathit{ds}_{edm}$ and $\mathit{ds}_{crm}$ in two cases: (1) only the top semantic type (the type with the highest confidence value) is considered (k=1), (2) the top four learned semantic types are taken into account as the candidate semantic types (k=4). Note that, when k=1, the MRR value is equal to the accuracy, i.e., how many of the attributes are labeled with their correct semantic types.

%\begin{figure}
%\begin{center}
%\scalebox{0.9} {\includegraphics[width=.95\linewidth]{figures/scenario2-edm-mrr}}
%\caption{MRR value of the learned semantic types for $ds_{edm}$ when only the top learned semantic types are considered (k=1); and the top four suggested types are considered (k=4)}
%\label{fig:scenario2-edm-mrr}
%\end{center}
%\end{figure}
%
%\begin{figure}
%\begin{center}
%\scalebox{0.9} {\includegraphics[width=.95\linewidth]{figures/scenario2-crm-mrr}}
%\caption{MRR value of the learned semantic types for $ds_{crm}$ when only the top learned semantic types are considered (k=1); and the top four suggested types are considered (k=4)}
%\label{fig:scenario2-crm-mrr}
%\end{center}
%\end{figure}

\begin{figure}
\centering
\begin{subfigure}[b]{.48\textwidth}
  \centering
  \includegraphics[width=1\linewidth]{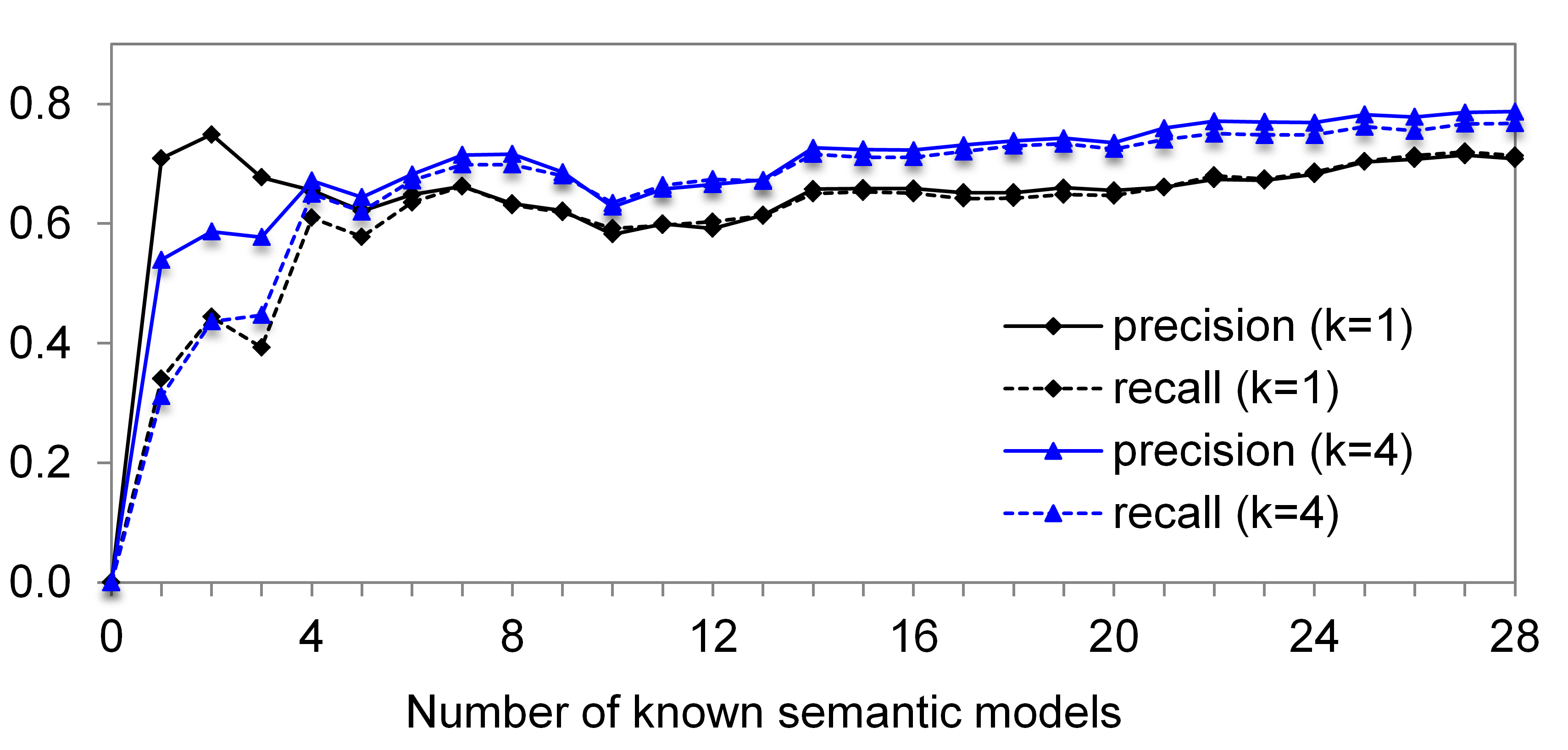}
  \caption{$\mathit{ds}_{edm}$}
  \label{fig:scenario2-edm-accuracy}
\end{subfigure} \\
% removing the new line (\\) from the end puts two images next to each other
\begin{subfigure}[b]{.48\textwidth}
  \centering
  \includegraphics[width=1\linewidth]{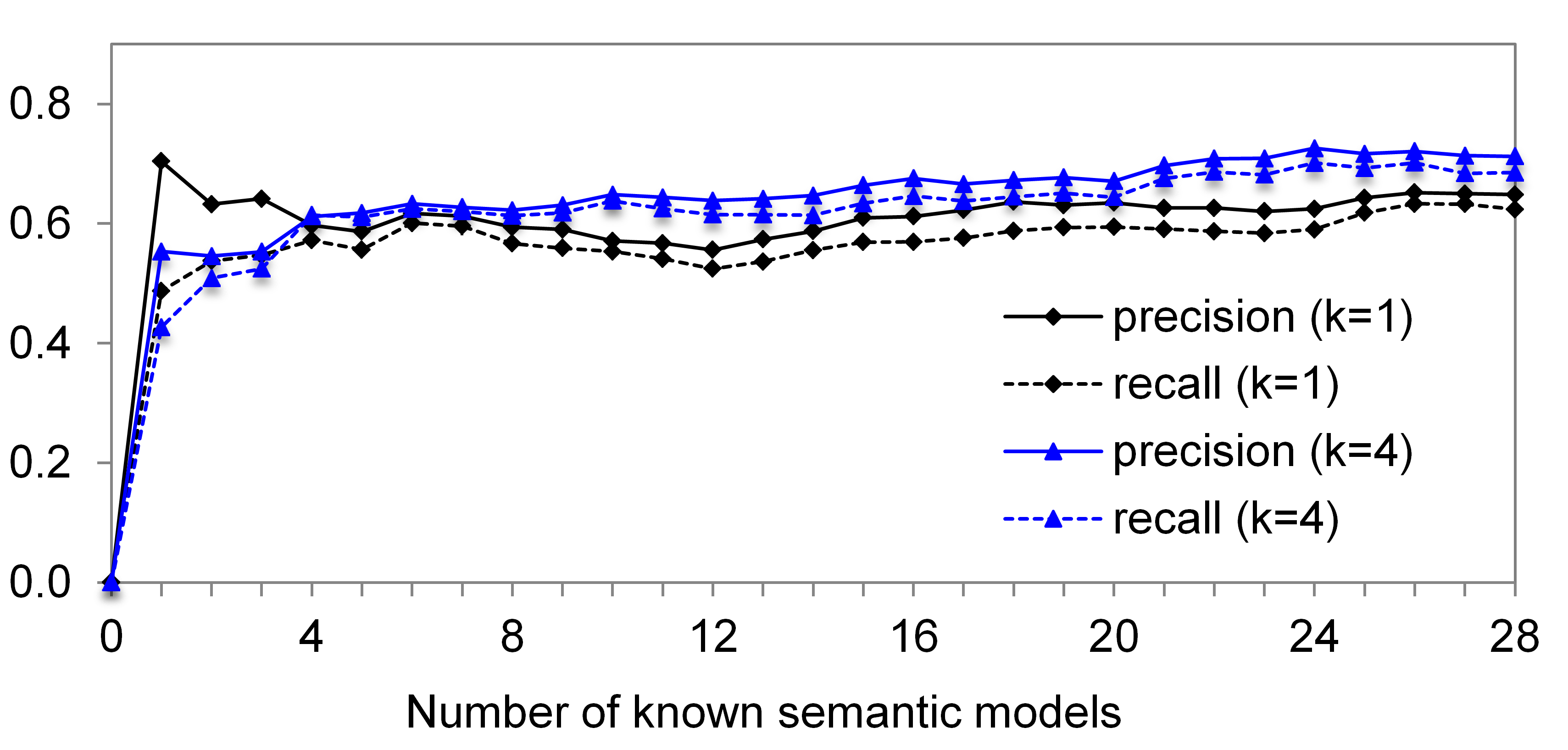}
  \caption{$\mathit{ds}_{crm}$}
  \label{fig:scenario2-crm-accuracy}
\end{subfigure}
\caption{Average precision and recall for the learned semantic models for k=1 and k=4.}
\vspace{0.05in}
\label{fig:scenario2-accuracy}
\end{figure}

Once the labeling is done, we feed the learned semantic types to the rest of algorithm to learn a semantic model for each source. The average precision and recall of the learned models are illustrated in Figure~\ref{fig:scenario2-accuracy}. The black color shows the precision and recall for k=1, and the blue color illustrates the precision and recall for k=4. In this experiment, we computed precision and recall for all the links including the links from the class nodes to the data nodes (these links are associated with the learned semantic types). The results show that using the known semantic models as background knowledge yields in a remarkable improvement in both precision and recall compared to the case in which we only consider the domain ontology (M\SB{0}). 

%\begin{figure}
%\begin{center}
%\scalebox{1} {\includegraphics[width=.95\linewidth]{figures/scenario2-edm-accuracy}}
%\caption{Average precision and recall for the learned semantic models for $ds_{edm}$ for k=1 and k=4}
%\label{fig:scenario2-edm-accuracy}
%\end{center}
%\end{figure}
%
%\begin{figure}
%\begin{center}
%\scalebox{1} {\includegraphics[width=.95\linewidth]{figures/scenario2-crm-accuracy}}
%\caption{Average precision and recall for the learned semantic models for $ds_{crm}$ for k=1 and k=4}
%\label{fig:scenario2-crm-accuracy}
%\end{center}
%\end{figure}

We provide an example to help in understanding the correlation between the MRR and the precision values, i.e., how the accuracy of the learned semantic types affects the accuracy of the learned semantic models. The average MRR value for $\mathit{ds}_{crm}$ when we use k=1 and M\SB{28} (leave-one-out setting) is 0.75 (Figure~\ref{fig:scenario2-crm-mrr}). This means that our labeling algorithm can learn the correct semantic types for only 75\% of the attributes. From Table~\ref{tab:datasets}, we know that the gold standard models for $\mathit{ds}_{crm}$ have totally 418 data nodes, and thus, 418 links in the gold standard models correspond to the source attributes. Since 75\% of the attributes are labeled correctly, 313 links out of 418 links corresponding to the source attributes will be correct in the learned semantic models. Even if we predict all the internal links correct (785-418=367 links), the maximum precision would be $86\%$ ((367+313)/785). However, the input to the Steiner tree algorithm are the nodes coming from the learned semantic types (leaves of the tree), and incorrect semantic types may prompt the Steiner tree algorithm to select incorrect links in the higher levels (internal links). As we see in Figure~\ref{fig:scenario2-crm-accuracy}, in the k=1 and M\SB{28} setting, the average precision of the learned semantic models is 65\%.

When considering the top four semantic types (k=4) instead of only the top one semantic type (k=1), our algorithm recovers some of the correct semantic types even if they are not the top predictions of the labeling function. For example, in the dataset $\mathit{ds}_{crm}$, using k=4 rather than k=1 when we have 28 known models (M\SB{28}), improves the precision by $6\%$ and the recall by $7\%$ (Figure~\ref{fig:scenario2-crm-accuracy}). This improvement is mainly because of the coherence factor we take into account in scoring the mappings and also ranking the candidate semantic models. 

\begin{figure}
\centering
\begin{subfigure}[b]{0.48\textwidth}
  \centering
  \includegraphics[width=1\linewidth]{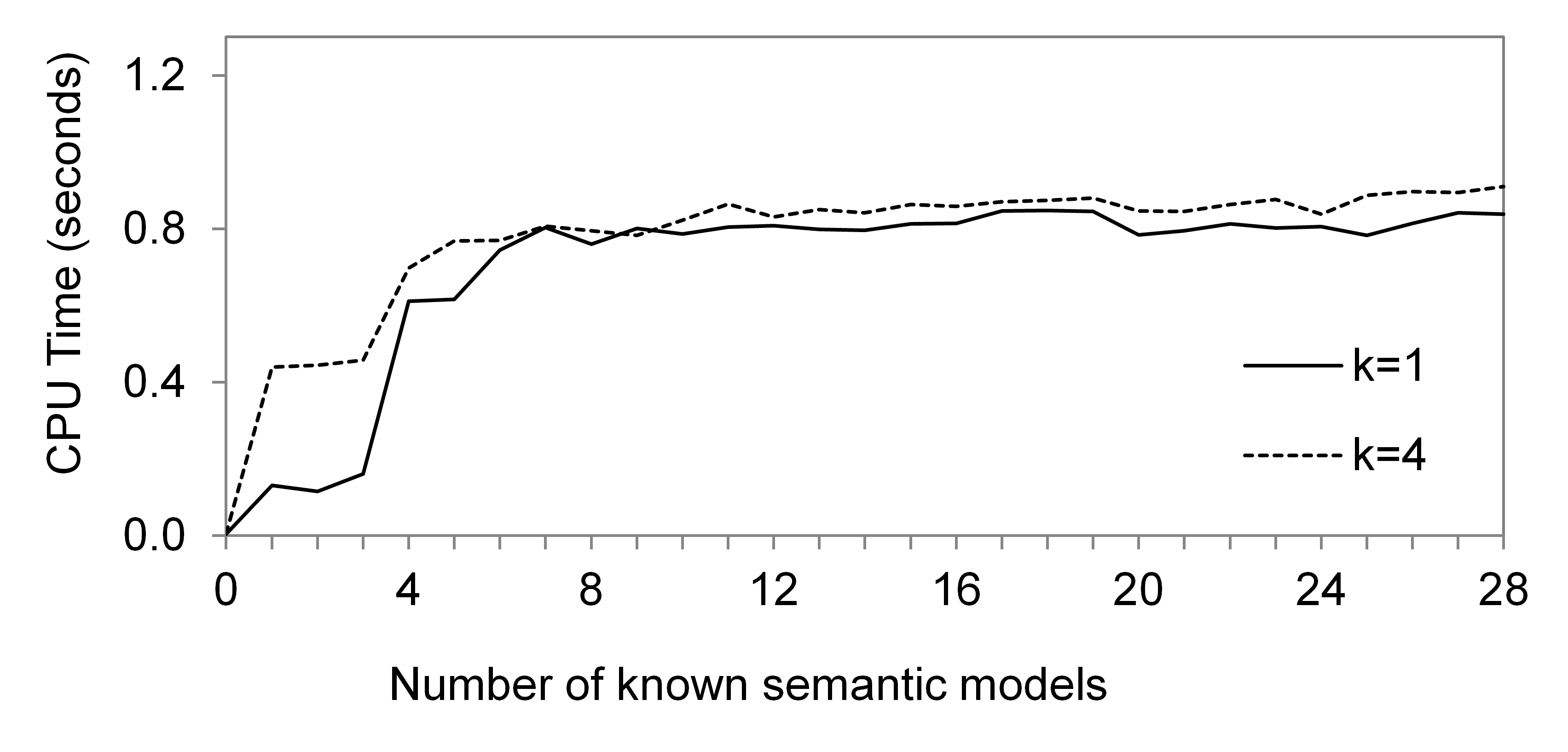}
  \caption{$\mathit{ds}_{edm}$}
  \label{fig:scenario2-edm-time}
\end{subfigure} \\
% removing the new line (\\) from the end puts two images next to each other
\begin{subfigure}[b]{0.48\textwidth}
  \centering
  \includegraphics[width=1\linewidth]{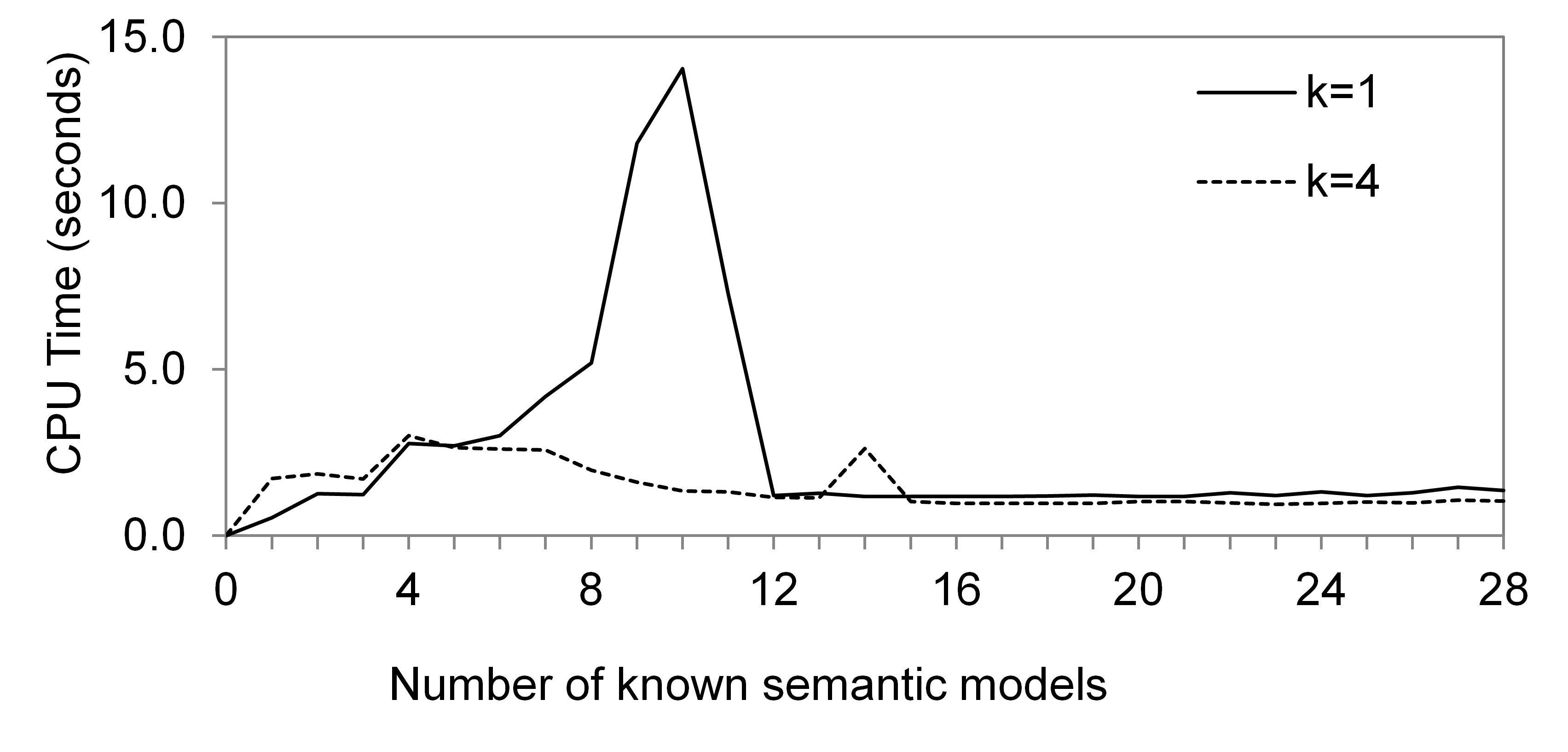}
  \caption{$\mathit{ds}_{crm}$}
  \label{fig:scenario2-crm-time}
\end{subfigure}
\caption{Average semantic model learning time when the attributes are labeled with their correct semantic types.}
\vspace{0.05in}
\label{fig:scenario2-time}
\end{figure}

The running time of the algorithm in the second scenario is displayed in Figure~\ref{fig:scenario2-time}. This time does not include the labeling step. The work done by Krishnamurthy et al. \cite{krishnamurthy2015} contains a detailed analysis of the performance of the labeling algorithm. As we can see in Figure~\ref{fig:scenario2-crm-time}, the running time of the algorithm is higher at M\SB{8}, M\SB{9}, M\SB{10}, and M\SB{11} when k=1. This is because computing top 10 Steiner trees takes longer once we add semantic models of $s_8$, $s_9$, $s_{10}$, and $s_{11}$ to the graph. When adding more semantic models, the algorithm runs faster. For example, the average time at M\SB{11} is 7.29 seconds while it is 1.21 seconds at M\SB{12}. This is the result of a combination of several reasons. First, there is more training data in learning the semantic types of a source $s_i$ at M\SB{12}, and this affects the output of the mapping algorithm (Algorithm~\ref{alg:mapping}). Second, the structure of the graph is different at M\SB{12} and this results in different mappings between the source attributes and the graph. Finally, the new semantic model $sm(s_{12})$ adds new paths to the graph allowing the Steiner tree algorithm to find the top 10 trees faster. 

%\begin{figure}
%\begin{center}
%\scalebox{1} {\includegraphics[width=.95\linewidth]{figures/scenario2-edm-time}}
%\caption{Average semantic model learning time for $ds_{edm}$ for k=1 and k=4}
%\label{fig:scenario2-edm-time}
%\end{center}
%\end{figure}
%
%\begin{figure}
%\begin{center}
%\scalebox{1} {\includegraphics[width=.95\linewidth]{figures/scenario2-crm-time}}
%\caption{Average semantic model learning time for $ds_{crm}$ for k=1 and k=4}
%\label{fig:scenario2-crm-time}
%\end{center}
%\end{figure}

\begin{figure}
\centering
\begin{subfigure}[b]{.48\textwidth}
  \centering
  \includegraphics[width=1\linewidth]{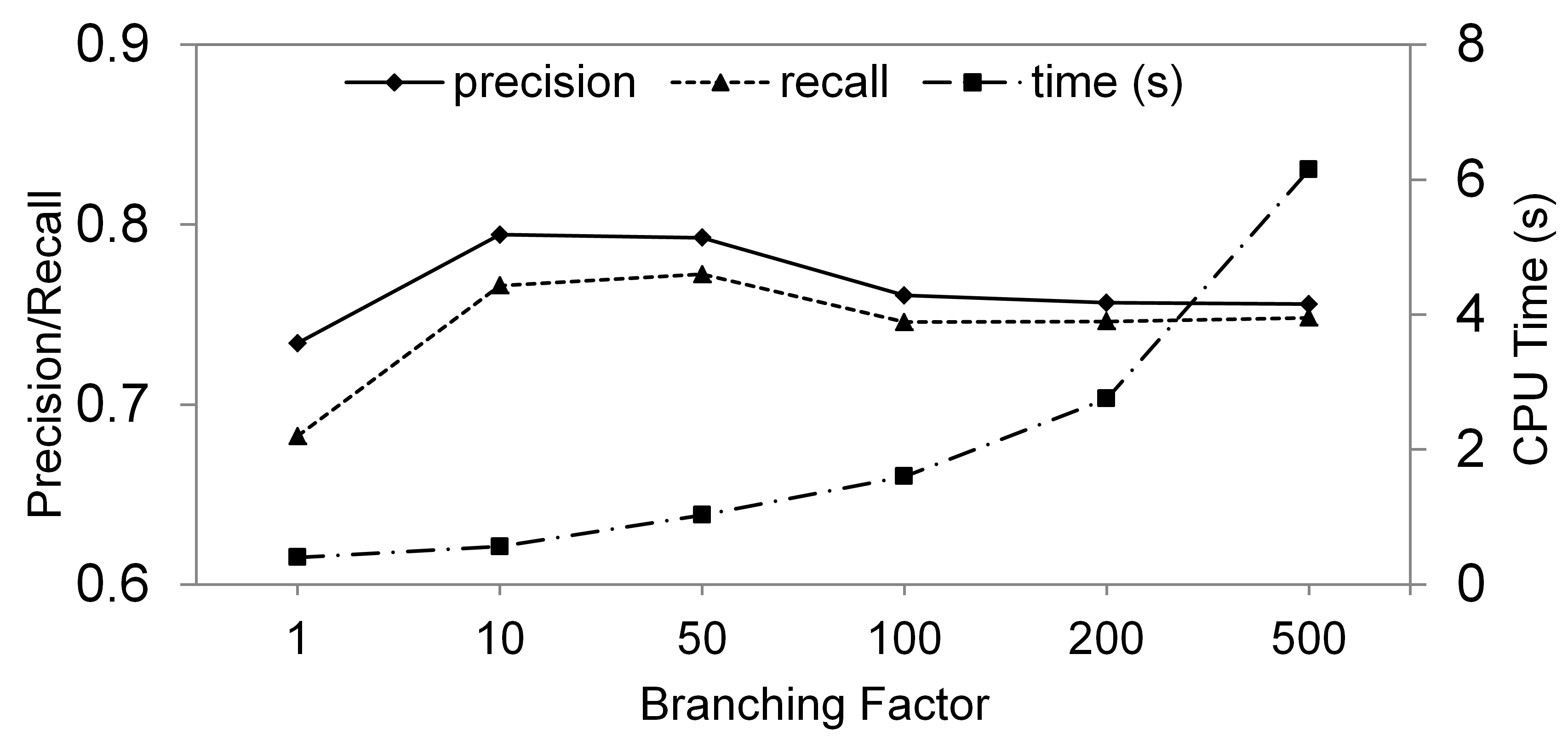}
  \caption{$\mathit{ds}_{edm}$}
  \label{fig:scenario2-edm-branching-factor}
\end{subfigure} \\
% removing the new line (\\) from the end puts two images next to each other
\begin{subfigure}[b]{.48\textwidth}
  \centering
  \includegraphics[width=1\linewidth]{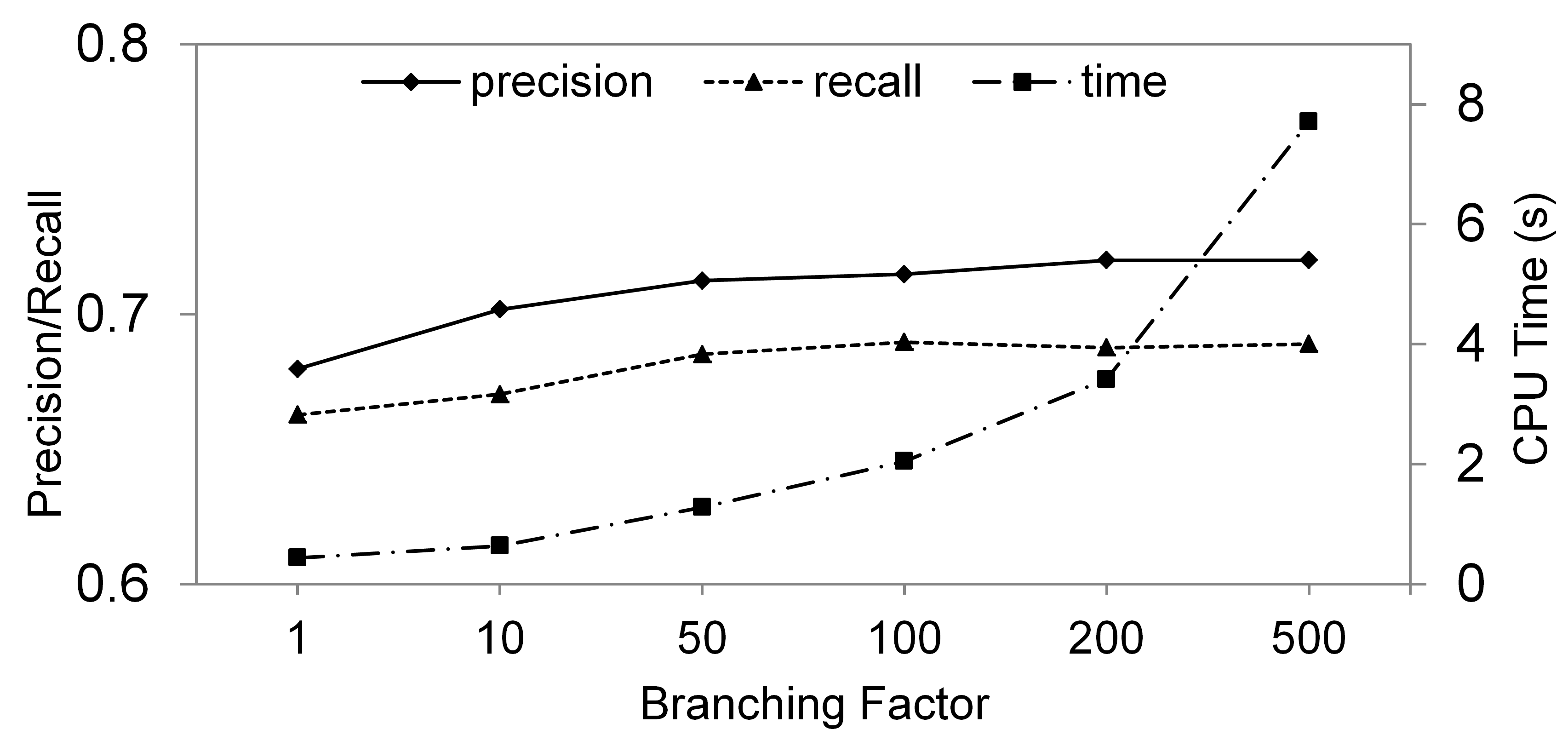}
  \caption{$\mathit{ds}_{crm}$}
  \label{fig:scenario2-crm-branching-factor}
\end{subfigure}
\caption{impact of branching factor on precision, recall, and running time for k=4 and M\SB{28}.}
\vspace{0.05in}
\label{fig:scenario2-branching-factor}
\end{figure}

We mentioned earlier that we used 50 as the value of the branching factor in  mapping the source attributes to the graph (line 21 of Algorithm~\ref{alg:mapping}). The branching factor is essential to the scalability of our mapping algorithm. This value can be configured by trying some sample values and then choosing a value yielding good accuracy while keeping the running time of the algorithm reasonably low. This can be different for each dataset. In our evaluation, using \textit{branching\U factor=50} worked well for both datasets. Figure~\ref{fig:scenario2-branching-factor} illustrates how changing the value of the branching factor affects the precision, recall, and running time of the algorithm in a setting where we considered 4 candidate semantic types (k=4) and the semantic models of all the other sources were known (M\SB{28}). In this experiment, we fixed the value of \textit{num\U of\U candidates} (line 27 of Algorithm~\ref{alg:mapping}) equal to the value of \textit{branching\U factor}. This means that all the generated mappings will be given to the Steiner tree algorithm as the candidate mappings. As we can see in Figure~\ref{fig:scenario2-crm-branching-factor}, increasing the value of the branching factor from 50 to 200 for $\mathit{ds}_{crm}$ provides $1\%$ improvement in the precision, however, it increases the average running time by 2.14 seconds. We chose to ignore this insignificant increase in the precision and used 50 as the branching factor to gain a better running time.

%\begin{figure}
%\begin{center}
%\scalebox{1} {\includegraphics[width=.95\linewidth]{figures/scenario2-edm-branching-factor}}
%\caption{Impact of branching factor on precision, recall, and running time for $ds_{edm}$  k=4 and M\SB{28}}
%\label{fig:scenario2-edm-branching-factor}
%\end{center}
%\end{figure}
%
%\begin{figure}
%\begin{center}
%\scalebox{1} {\includegraphics[width=.95\linewidth]{figures/scenario2-crm-branching-factor}}
%\caption{Impact of branching factor on precision, recall, and running time for $ds_{crm}$ for k=4 and M\SB{28}}
%\label{fig:scenario2-crm-branching-factor}
%\end{center}
%\end{figure}

\section{Related Work}
\label{sec:related}

The problem of describing semantics of data sources is at the core of data integration \cite{doan2012} and exchange \cite{arenas2010}. The main approach to reconcile the semantic heterogeneity among sources consists of defining logical mappings between the source schemas and 
a common target schema. One way to define these mappings is local-as-view (LAV) descriptions where every source is defined as a view over the domain schema \cite{doan2012}. The semantic models that we generate are graphical representation of LAV rules, where the domain schema is the domain ontology.
Although the logical mappings are declarative,
%  and often concise, 
defining them requires significant technical expertise, so there has been much interest in techniques that facilitate their generation. 

In traditional data integration, the mapping generation problem is usually decomposed in a \textit{schema matching} phase followed by \textit{schema mapping} phase \cite{bellahsene2011}. Schema matching \cite{rahm01} finds correspondences between elements of the source and target schemas. For example, iMAP \cite{dhamankar04} discovers complex correspondences by using a set of special-purpose searchers, ranging from data overlap, to machine learning and equation discovery techniques. 
This is analogous to the semantic labeling step in our work \cite{krishnamurthy2015}, where we learn a labeling function to learn candidate semantic types for a source attribute. Every semantic type maps an attribute to an element in the domain ontology (a class or property in the domain ontology). 

Schema mapping defines an appropriate transformation that populates the target schema with data from the sources. Mappings may be arbitrary procedures, but of greater interest are declarative mappings expressible as queries in SQL, XQuery, or Datalog.
% , and more specifically as st-tgds. 
These mapping formulas are generated by taking into account the schema matches and schema constraints. 
There has been much research in schema mapping, from the seminal work on Clio \cite{fagin2009:clio}, which provided a practical system and furthered the theoretical foundations of data exchange \cite{fagin2005} to more recent systems that support additional schema constraints  \cite{marnette2011}.
%
% that generate mappings supporting more complex schema constraints in a scalable way \cite{marnette2010}.  
%
Alexe et al. \cite{alexe2011} generate schema mappings from examples of source data tuples and the corresponding tuples over the target schema. An et al. \cite{an2007} generate declarative mapping expressions between two tables with different schemas starting from element correspondences. They create a graph from the conceptual model (CM) of each schema and then suggest plausible mappings by exploring low-cost Steiner trees that connect those nodes in the CM graph that have attributes participating in element correspondences. Their work is similar to our previous  semi-automatic approach to build the semantic models \cite{knoblock12:semi}, where we derive a graph from the domain ontology and the learned semantic types. We exploited the knowledge from the ontology to assign weights to the links based on their types, e.g., direct properties get lower weight than inherited properties, because we wanted to give more priority to more specific relations. We also allow the user to correct the mappings interactively. In the current paper, in addition to the ontology, we consider previous known semantic models to improve the modeling of an unknown source.

Our work on learning semantic models of structured sources is complementary to these schema mapping techniques. Instead of focusing on satisfying schema constraints, we analyze known source models to propose mappings that capture more closely the semantics of the target source in ways that schema constraints could not disambiguate.  For example, by suggesting that a \textit{dcterms:creator} relationship is more likely than \textit{dbpedia:owner} in a given domain. Moreover, our algorithm can incrementally refine the mappings based on user feedback and learn from this feedback to improve future predictions.

%In the traditional data integration,  source descriptions are specified as global-as-view (GAV) or local-as-view (LAV) descriptions.
%There has been much work to automate the task of generating these mappings.
%Some work \cite{rahm01,dhamankar04} only finds correspondences between elements of the source and global schemas. This is analogous to the semantic labeling step in our work, where we use a machine learning technique \cite{goel12:exploiting} to learn candidate semantic types for a source attribute. Every semantic type maps an attribute to an element in the domain ontology (a class or property in the domain ontology). 
%
%Further work \cite{bellahsene2011,fagin2009:clio,alexe2011,an2007} generates more complex mappings that express relationships. However, they do not exploit the knowledge of previously modeled sources as we do.

In the Semantic Web, what is meant by a source description is a semantic model describing the source in terms of the concepts and relationships defined by a domain ontology. There are many studies on mapping data sources to ontologies. Several approaches
have been proposed to generate semantic web data from databases and spreadsheets \cite{sahoo2009:survey}.

D2R \cite{Bizer03,bizer2006d2r} and D2RQ \cite{Bizer04} are mapping languages that enable the user to define mapping rules between tables of relational databases and target ontologies in order to publish semantic data in RDF format. R2RML \cite{R2RML} is a another mapping language, which is a W3C recommendation for expressing customized mappings from relational databases to RDF datasets. Writing the mapping rules by hand is a tedious task. The users need to understand how the source table maps to the target ontology. They also need to learn the syntax of writing the mapping rules. 
RDOTE \cite{vavliakis2010:rdote} is a tool that provides a graphical user interface to facilitate mapping relational databases into ontologies. The developers of RDOTE have said they will incorporate an export/import mechanism for D2RQ compliant mapping files, as well as a query builder graphical user interface to hasten the mapping creation process. RDF123 \cite{han2008:rdf123} and XLWrap \cite{langegger2009:xlwrap} are other tools to define mappings from spreadsheets to RDF graphs. Although these tools can facilitate the mapping process, the users still need to manually define the mappings between the source and target ontologies.

%\cite{ding2010:twc}: generate rdf from government data

In recent years, there are some efforts to automatically infer the implicit semantics of tables. Polfliet and Ichise \cite{polfliet2010:automated} use string similarity between the column names and the names of the properties in the ontology to find a mapping between the table columns and the ontology. Wang et al. \cite{wang2012:understanding} detect the header of Web tables  and use them along with the values of the rows to map the columns to the attributes of the corresponding entity in a rich and general purpose taxonomy of worldly facts built from a corpus of over one million Web pages and other data. This approach can only deal with the tables containing information of a single entity type.

Limaye et al. \cite{limaye2010} used YAGO\footnote{\url{http://www.mpi-inf.mpg.de/yago-naga/yago}} to annotate web tables and generate binary relationships using machine learning approaches. However, this approach is limited to the labels and relations defined in the YAGO ontology (less than 100 binary relationships). Venetis et al. \cite{venetis2011} presented a scalable approach to describe the semantics of tables on the Web. To recover the semantics of tables, they leverage a database of class labels and relationships automatically extracted from the Web. They attach a class label to a column if a sufficient number of the values in the column are identified with that label in the database of class labels, and analogously for binary relationships. Although these approaches are very useful in publishing semantic data from tables, they are limited in learning the semantics relations. Both of these approaches only infer individual binary relationships between pair of columns. They are not able to find the relation between two columns if there is no direct relationship between the values of those columns. Our approach can connect one column to another one through a path in the ontology. For example, suppose that we have a table including two columns \textit{person} and \textit{city}, where the city is the location of the company the person is working for. Our approach can learn a semantic model that connects the class \textit{Person} to the class \textit{City} through the chain $Person \overset{worksFor}{\longrightarrow} Organization \overset{location}{\longrightarrow} City$. 

There is also work that exploits the data available in the Linked Open Data (LOD) cloud to capture the semantics of the tables and publish their data as RDF. Munoz et al.\ \cite{munoz13:triplifying} mine RDF triples from the Wikipedia tables by linking the cell values to the resources available in DBPedia \cite{Auer07:dbpedia}. This approach is limited to Wikipedia tables because of its simple linking algorithm. If a cell value contains a hyperlink to a Wikipedia page, the Wikipedia URL maps to a DBpedia entity URI by replacing the namespace \url{http://en.wikipedia.org/wiki/} of the URL with \url{http://dbpedia.org/resource/}. 

In other work, Mulwad et al. \cite{mulwad2013semantic} used \textit{Wikitology} \cite{syed2011:creating}, an ontology which combines some existing manually built knowledge systems such as DBPedia and Freebase \cite{Bollacker08:freebase}, to link cells in a table to Wikipedia entities. They query the background LOD to generate initial lists of candidate classes for column headers and cell values and candidate properties for relations between columns. Then, they use a probabilistic graphical model to find the correlation between the columns headers, cell values, and relation assignments. The quality of the semantic data generated by this category of work is highly dependent to how well the data can be linked to the entities in LOD. While for most popular named entities there are good matches in LOD, many tables contain domain-specific information or numeric values (e.g., temperature and age) that cannot be linked to LOD. Moreover, these approaches are only able to identify individual binary relationships between the columns of a table. However, an integrated semantic model is more than fragments of binary relationships between the columns. In a complete semantic model, the columns may be connected through a path including the nodes that do not correspond to any column in the table.

%\cite{sekhavat14:knowledge}: a probabilistic model using patterns extracted from Web (NLP) to convert table data to RDF - binary relationships
%

%Semantic annotation of services have also received attention. Annotating the input and output parameters of Web services and Web APIs is useful for automatic service discovery and composition. SAWSDL vocabulary \cite{farrell07:semantic} allows adding semantic meta-data to service descriptions. Service inputs and outputs can be annotated by the concepts and properties using an attribute called \textit{modelReference}. SWEET \cite{maleshkova09:sweet} is a tool that supports users in creating semantic descriptions of RESTful services. There has been some work on classifying Web services into different domains \cite{Hess03} and automatically labeling the input and outputs of Web services \cite{lerman06,saquicela2011}. This work provides useful knowledge for service discovery, but not sufficient for automating service integration. In our work we are able to learn more expressive descriptions of Web services that describe how the attributes of a service relate to one another.

%Saquicela et al. \cite{saquicela2011} presented a technique that automatically generates a lightweight semantic annotation of geospatial services. They rely on the names of the service parameters and their synonyms to find mappings from the service parameters to the classes and properties of DBPedia and GeoNames ontologies. This approach does not extract the relationships between the service inputs and outputs.

Parundekar et al.\ \cite{parundekar2012:iswc} previously developed an approach to automatically generate conjunctive and disjunctive mappings between the ontologies of linked data sources by exploiting existing linked data instances. However, the system does not model arbitrary sources such as we present in this paper. Carman and Knoblock \cite{carman07} use known source descriptions to learn a semantic description that precisely describes the relationship between the inputs and outputs of a source, expressed as a Datalog rule. However, their approach is limited in that it can only learn sources whose models are subsumed by the models of known sources. That is, the description of a new source is a conjunctive \textit{combination} of known source descriptions. By exploring paths in the domain ontology, in addition to patterns in the known sources, we can hypothesize target mappings that are more general than previous source descriptions or their combinations.

In our earlier Karma work \cite{knoblock12:semi}, we build a graph from learned semantic types and a domain ontology and use this graph to map a source to the ontology {\em interactively}. In that work, the system uses the knowledge from the domain ontology to propose models to the user, who can correct them as needed. The system remembers semantic type labels assigned by the user, however, it does not learn from the structure of previously modeled sources.

%Most of the related work is manual, or automates learning semantic types for table columns, but is limited in learning relationships. 

% In our previous work \cite{taheriyan13:iswc,taheriyan14:icsc}, we presented an automatic approach that exploits the known semantic models to learn a semantic model for a new unknown source. We analyze known semantic models to hypothesize semantic models that capture more closely the semantics of the new source. When there are many source attributes, there will be a large number of mappings from the source attributes to the nodes of the graph. This approach, uses a beam search algorithm in the mapping step to overcome this problem. However, the graph grows as the number of known semantic models grows, which makes computing the semantic models inefficient.

The most closely related work \cite{taheriyan13:iswc,taheriyan14:icsc} on exploiting known semantic models to learn a model for a new unknown source. However, our previous approach was less scalable.  When there are many source attributes, there will be a large number of mappings from the source attributes to the nodes of the graph. Even though we used a beam search algorithm in the mapping step to ameliorate this problem, the graph grows as the number of known semantic models grows, which makes computing the semantic models inefficient. 
In this paper, we have presented a compact graph structure that merges overlapping segments of the known semantic models. We also use a new algorithm to generate and rank the candidate semantic models. We generate candidate models by computing \textit{top-k} Steiner trees and then rank them based on the coherence of the links. This new approach, in addition to generating more accurate semantic models, significantly improves the running time of the learning process. Integrating our algorithm into Karma, enables the user to refine the automatically learned models resulting in more accurate predictions for future data sources.

In recent years, ontology matching has received much attention in the Semantic Web community \cite{kalfoglou:2003,pavel:2013}. Ontology matching (or ontology alignment) finds the correspondence between semantically related entities of different ontologies. This problem is analogous to schema matching in databases. Both schemas and ontologies provide a vocabulary of terms that describe a domain of interest. However, schemas often do not provide explicit semantics for their data. 
Our work benefits from some of the techniques developed for ontology matching. For example, \textit{instance-based ontology matching} exploits similarities between instances of ontologies in the matching process. Our semantic labeling algorithm adopts the same idea to map the data of a new source to the classes and properties of a target ontology. The algorithm computes the similarity (cosine similarity between TF/IDF vectors) between the data of the new source and the data of the sources whose semantic models are known. 

Ontology matching is different than the problem we addressed in this paper in the sense that in our work the data that is being mapped to a target ontology is not bound to any source ontology. This makes our problem more complicated since no explicit semantics is necessarily attached to data sources. Moreover, most of the work on ontology matching only finds simple correspondences such as equivalence and subsumption between ontology classes and properties. Therefore, the explicit relationships within the data elements are often missed in aligning the source data to the target ontology. Suppose that we want to find the correspondences between a source ontology $O_s$ and a target ontology $O_t$. Using ontology matching, we find that the class $A_s$ in $O_s$ maps to the class $A_t$ in $O_t$ and the class $B_s$ in $O_s$ maps to the class $B_t$ in $O_t$. Assume that there is only one property connecting $A_s$ to $B_s$ in $O_s$, but there are multiple paths connecting $A_t$ to $B_t$ in $O_t$. If we align the source data to the target ontology $O_t$ using the correspondences found by ontology matching, the instances of $A_s$ will be mapped to the class $A_t$ and the instances of $B_s$ will be mapped to the class $B_t$. However, this alignment does not tell us which path in $O_t$ captures the correct meaning of the source data.

\section{Discussion}
\label{sec:discussion}

% We presented a scalable approach to learn semantic models of structured data sources as a mapping from the source to a domain ontology. The idea is to exploit the knowledge of previously learned semantic models to hypothesize a plausible semantic model for a new source. This work builds our previous work on learning semantic models \cite{taheriyan13:iswc} and addresses the uncertainty in learning semantic types and provides an approach that scales to sources with many attributes. Our evaluation shows that the new approach can learn semantic models for data sources that the previous approach was not able to handle.
% %Our evaluation shows that the new approach can learn more accurate semantic models for data sources. It can also learn semantic models for data sources that the previous approach was not able to deal with them. 

In this paper, we presented a scalable approach to learn semantic models of structured data sources as mappings from the sources to a domain ontology. Such models are the key ingredients in the process of publishing data
% sources 
into the LOD knowledge graph. The core idea is to exploit %the knowledge from 
the domain ontology and previously learned semantic models to hypothesize a plausible semantic model for a new source. The evaluation shows that our approach learns rich semantic models with minimal user input. 

%\hl{Comment from Pedro:The term ``learning semantic models'' in this context is confusing. Here, it means ``hypothesizing  a plausible model''. Some learning happens when user label columns with semantic types or when they publish models. I would use the term ``inferring a semantic model'' or ``hypothesizing'' as you used in the previous sentence.}

The first step in learning semantic models is learning the semantic types in which the system labels each source attribute with a class or property from the ontology. The output of the labeling step is a set of candidate semantic types and their confidence values rather than one fixed semantic type. Taking into account the uncertainty of the labeling algorithm is very important because machine learning techniques often cannot distinguish the types of the source attributes that have similar data values, e.g., \textit{birthDate} and \textit{deathDate}.

Once the system produces candidate semantic types for each attribute, it creates a graph from known semantic models and augments it by adding the nodes and the links corresponding to the semantic types and adding the paths inferred from the ontology. The next step is mapping the source attributes to the nodes of the graph where we use a search algorithm that enables the system to do the mapping even when the source has many attributes. The algorithm, after processing each source attribute, prunes the existing mappings by scoring them and removing the ones having lower scores. The proposed scoring function not only contributes to the scalability of our method, but also increases the accuracy of the learned models. 

The final part of the approach is computing the minimal tree that connects the nodes of the candidate mappings. This step might be computationally inefficient if we have a very large graph. However, our algorithm to construct the graph consolidates the overlapping segments of the known semantic models, making it scalable to a huge number of known semantic models. 

Our learning algorithms play an important role in making the Karma interactive user interface easy to use, a key design goal given that many of our users are domain experts, but are not Semantic Web experts.
Our experience observing users is that they can understand and critique models when displayed in our interactive user interface.
They can easily verify that models accurately capture the semantics of a source, and can easily spot errors or controversial modeling decisions.
Users can click on the corresponding elements on the screen and do local modifications such as replacing the property of a link or changing the source or destination of a link.

We also observe that it is much harder for users to model a source from scratch, as is necessary in tools such as Open Refine.\footnote{\url{http://openrefine.org/}} Even though the user interface is easy to use, the task of filling a blank page with a model is daunting for many users.
Karma helps these users because it gives them an almost-correct model as a starting point.
Users can easily find the elements they do not agree with, and can easily change them.
A possible direction for future work is to perform user evaluations to measure the quality of the models produced using learning algorithms.
Although time to create models is important, we hypothesize that most users, such as our museum users, are primarily concerned with producing correct models, and time to model is a secondary concern for them. By using previous models, users are more likely to model sources in a correct way.

Our work also plays a role in helping communities to produce consistent Linked Data so that sources containing the same type of data use the same classes and properties when published in RDF.
Often, there are multiple correct ways to model the same type of data.
For example, users can use Dublin Core and FOAF to model the creator relationship between a person and an object (\textit{dcterms:creator} and \textit{foaf:maker}).
A community is better served when all the data with the same semantics is modeled using the same classes and properties.
Our work encourages consistency because our learning algorithms bias the selection of classes and properties towards those used more frequently in existing models.

A future direction of our work is to improve the quality of the automatically generated models by leveraging the significant amount of data available in the Linked Open Data (LOD) cloud, which is a vast and growing collection of semantic data that has been published by various data providers. The current estimate is that the LOD cloud contains over 30 billion RDF triples. Even the New York Times is now publishing all of their metadata as Linked Open Data.\footnote{See \url{http://data.nytimes.com}} It should be noted that a nontrivial portion of LOD is just data with limited semantic descriptions, but much of that data has been linked to other sources that does have some form of semantic description. Given the growing availability of this type of data, LOD will provide an invaluable source of semantic content that we can exploit as background knowledge.

Given the huge repository of data available in LOD, for any given set of values provided by a new source, we can search for classes that provide or even subsume all of the data for a given property of a source. For example, if we have a set of values for people names or temperature, we are likely to find some classes in LOD that provides that same set of values. We will not require a perfect overlap between the set of values from the source and a class in the Linked Open Data, but rather a statistically significant overlap, similar to what is done by Parundekar et al.\ \cite{parundekar2012:iswc}. An important challenge here is how to efficiently find the classes that most closely match the set of attribute values and how to handle the problem that the classes that match the best may come form different ontologies. 

We can also exploit LOD to disambiguate the relationships between the attributes \cite{taheriyan15:cold}. Once we have identified the semantic types of the source attributes, we can search for corresponding classes in LOD and analyze which properties are connecting them. Those properties can be candidates for the relationships between the attributes of the new source. Consider the semantic model of the source $dia$ in Figure~\ref{fig:dia}. Once we identify that \textit{$\langle$aac:CulturalHeritageObject,dcterms:title$\rangle$} and \textit{$\langle$aac:Person,foaf:name$\rangle$} are the semantic types of the first and fourth attributes, we can search LOD for possible properties between instances of the classes \textit{aac:CulturalHeritageObject} and \textit{aac:Person} and find that the properties \textit{dcterms:creator} and \textit{acc:sitter} are better candidates than other properties that ontology suggests, e.g., \textit{dbpedia:owner}. By combining the information we extract for each pair of classes, we can narrow the search to those classes and properties that commonly occur together.

\section*{Acknowledgements}
This research was supported in part by the National Science Foundation under Grant No. 1117913 and in part by Defense Advanced Research Projects Agency (DARPA) via AFRL contract numbers FA8750-14-C-0240 and FA8750-16-C-0045. The U.S. Government is authorized to reproduce and distribute reprints for Governmental purposes notwithstanding any copyright annotation thereon. The views and conclusions contained herein are those of the authors and should not be interpreted as necessarily representing the official policies or endorsements, either expressed or implied, of NSF, DARPA, AFRL, or the U.S. Government. We would like to thank the anonymous reviewers for their valuable comments and suggestions to improve the paper. We are also grateful to Yinyi Chen for her help in creating the gold standard models for our evaluation.



%% References

%\nocite{*} % prints all the items in the bib file

%\bibliographystyle{elsarticle-num}
\bibliographystyle{elsarticle-num-modified}
\biboptions{sort&compress}
\bibliography{references}






% \end{thebibliography}

\end{document}